\pdfoutput=1

\documentclass[11pt]{article}

\usepackage[final]{coling}

\usepackage{times}
\usepackage{latexsym}
\usepackage{amsmath, amssymb}
\usepackage{dsfont}
\usepackage[T1]{fontenc}

\usepackage[utf8]{inputenc}
\usepackage{makecell, supertabular}
\usepackage{booktabs}
\usepackage{multirow}

\usepackage{xcolor}
\definecolor{darkgreen}{rgb}{0.0, 0.5, 0.0}
\definecolor{darkblue}{rgb}{0.0, 0.0, 0.75}
\definecolor{darkred}{rgb}{0.75, 0.0, 0.0}

\usepackage{microtype}
\usepackage{supertabular}

\usepackage{fancyhdr}

\usepackage{inconsolata}

\usepackage{graphicx}
\usepackage{caption}
\usepackage{subcaption}

\DeclareFontFamily{U}{matha}{\hyphenchar\font45}
\DeclareFontShape{U}{matha}{m}{n}{
      <5> <6> <7> <8> <9> <10> gen * matha
      <10.95> matha10 <12> <14.4> <17.28> <20.74> <24.88> matha12
      }{}
\DeclareSymbolFont{matha}{U}{matha}{m}{n}

\DeclareMathSymbol{\Lt}{3}{matha}{"CE}
\DeclareMathSymbol{\Gt}{3}{matha}{"CF}

\newcommand\methodname{\textcolor{black}{\textsc{URIEL+}}}
\newcommand\uriel[1][]{\textcolor{black}{\textsc{URIEL#1}}}
\newcommand\langtovec{\textcolor{black}{\texttt{lang2vec}}}

\newcommand\langrank{\textcolor{black}{\textsc{LangRank}}}
\newcommand\lingualchemy{\textcolor{black}{\textsc{LinguAlchemy}}}
\newcommand\proxylm{\textcolor{black}{\textsc{ProxyLM}}}
\DeclareMathOperator{\indicator}{\mathds{1}}

\title{URIEL+: Enhancing Linguistic Inclusion and Usability in a Typological and Multilingual Knowledge Base}

\author{\textbf{Aditya Khan}$^{\dagger*}$, \textbf{Mason Shipton}$^{\ddagger}$\thanks{\text{ }The authors contributed equally.}, \textbf{David Anugraha}$^{\dagger}$, \textbf{Kaiyao Duan}$^{\dagger}$, \\ \textbf{Phuong H. Hoang}$^{\dagger}$, \textbf{Eric Khiu}$^{\S}$, \textbf{A. Seza Doğruöz}$^{\&}$, \textbf{En-Shiun Annie Lee}$^{\dagger \ddagger}$ \\
  $^{\dagger}$University of Toronto, Canada \quad $^{\ddagger}$Ontario Tech University, Canada \\
  $^{\S}$University of Michigan, USA \quad $^{\&}$LT3, IDLab, Universiteit Gent, Belgium\\
\texttt{\{adityakhan,anugraha\}@cs.toronto.edu,  masonshipton25@gmail.com} \\ \texttt{as.dogruoz@ugent.be, annie.lee@ontariotechu.ca}}

\usepackage{xcolor}

\newif\ifshowcomments
\showcommentstrue 

\usepackage{svg}

\usepackage{pgfplots}
\usepackage{pgfplotstable}
\usepackage{pifont}

\pgfplotsset{compat=1.18}
\usepgfplotslibrary{groupplots}

\makeatletter
\newcommand\resetstackedplots{
\makeatletter
\pgfplots@stacked@isfirstplottrue
\makeatother
\addplot [forget plot,draw=none] coordinates{(0,0.8) (0,3.2) (0,5.8) (0,8.2) (0,10.8) (0,13.2) (0,15.8) (0,18.2)};
}
\makeatother

\begin{document}
\maketitle
\begin{abstract}

\begin{figure*}[!th]
  \includegraphics[width=1.0\textwidth]{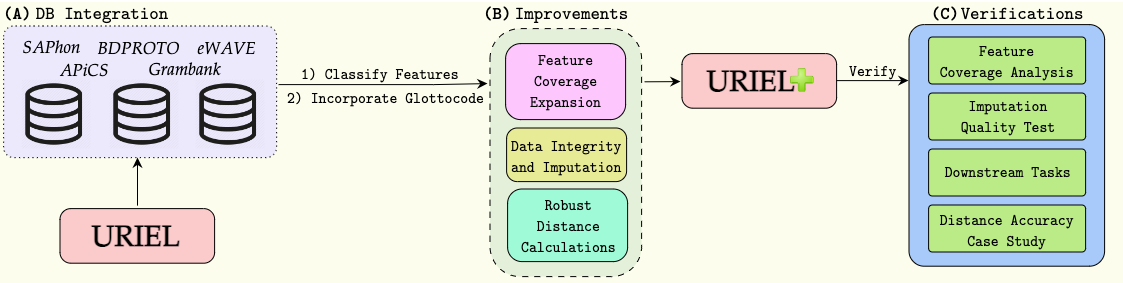}
  \caption{Overview of \uriel{+}. \textbf{(A)} We integrated five databases to incorporate more linguistic features across multiple languages, and \textbf{(B)} made several improvements on top of current \uriel{}. \textbf{(C)} We later verified the quality of \uriel{+} through various experiments.}
  \label{fig:overview}
\end{figure*}

\uriel{} is a knowledge base offering geographical, phylogenetic, and typological vector representations for $7970$ languages. It includes distance measures between these vectors for $4005$ languages, which are accessible via the \langtovec{} tool. Despite being frequently cited, \uriel{} is limited in terms of linguistic inclusion and overall usability. 
To tackle these challenges, we introduce \uriel{+}, an enhanced version of \uriel{} and \langtovec{} that addresses these limitations. In addition to expanding typological feature coverage for $2898$ languages, \uriel{+} improves the user experience with robust, customizable distance calculations to better suit the needs of users. These upgrades also offer competitive performance on downstream tasks and provide distances that better align with linguistic distance studies. The code is available at \url{https://github.com/Masonshipton25/URIELPlus}.

\end{abstract}

\section{Introduction}
\label{sec:intro}


The \uriel{} knowledge base and its \langtovec{} query tool \citep{littell-etal-2017-uriel} provide a standardized approach to representing languages as geographical, phylogenetic, and typological vectors. Geographical vectors contain distances between the locations where languages are spoken and $299$ latitude/longitude coordinates. Phylogenetic and typological vectors consist of binary indicators that denote membership in language families or structural features, respectively. \uriel{} enables language comparisons through \emph{language distance} calculations, which are performed using mathematical operations on these vectors.

\emph{Typological distance}, or differences in language structure, is foundational for cross-linguistic comparisons \citep{haspelmath-2023-word-class-universals} and plays a crucial role in multilingualism, second language acquisition, and natural language processing (NLP) \citep{NelsonMultilingual2021, haspelmath-2020-structural-uniqueness}. The challenge in defining \emph{typological distance} lies in the structural uniqueness of each language, which complicates direct comparisons \citep{haspelmath-2020-structural-uniqueness}. To navigate this complexity, linguists measure \emph{typological distance} by focusing on particular linguistic domains (e.g., syntax, phonology, or phonemic inventory) \cite{nerbonne-hinrichs-2006-linguistic}. 


\paragraph{Syntactic Distance}
Syntactic distance measures similarities and differences in grammatical structures using frameworks such as dependency trees and part-of-speech distributions. These methods provide quantitative comparisons of syntactic patterns between languages \cite{hammarstrom-oconnor-2013-dependency}.

\paragraph{Phonological Distance}
Phonological distance measures similarities and differences in the overall sound systems of languages, including both segmental and suprasegmental features. This involves analyzing phonetic properties like voicing and place of articulation, as well as prosodic elements such as stress and intonation. Tools like $n$-gram models and phoneme frequency analysis position languages in a multidimensional space based on these comprehensive phonological characteristics \cite{gamallo-pichel-alegria-2017-language-id}.

\paragraph{Phonemic Inventory Distance}  
Phonemic inventory distance measures the similarities and differences between the sets of phonemes in two languages, including both vowels and consonants. This involves comparing the number and types of phonemes present in each language, as well as their specific combinations, thereby providing insights into their phonemic structure to quantify how similar or different the phonemic inventories of the languages are \cite{BRADLOW2010930}.

\begin{table}[!th]
    \centering
    \resizebox{\columnwidth}{!}{%
    \begin{tabular}{lllllll}
        \toprule
        \textbf{Language Pair} & \textbf{geo} & \textbf{gen} & \textbf{fea} & \textbf{syn} & \textbf{pho} & \textbf{inv} \\ 
        \midrule
        \multirow{2}{*}{English-French} &  0.00 & 0.90 & 0.50 & 0.46 & 0.43 & 0.48 \\ 
        & \textbf{0.02} & \textbf{0.94} & \textbf{0.48} & \textbf{0.45} & \textbf{0.39} & \textbf{0.48} \\
        \midrule
        \multirow{2}{*}{Croatian-Serbian} & 0.00 & 0.13 & 0.90 & 0.71 & 0.83 & 0.68 \\
        & \textbf{0.02} & \textbf{0.00} & \textbf{0.00} & \textbf{0.00} & \textbf{NA} & \textbf{NA} \\
        \bottomrule
    \end{tabular}
    }
    \caption{Geographical (geo), genetic (gen), featural (typological) (fea), syntactic (syn), phonological (pho), and inventory (inv) distances for English-French in \uriel{} and \uriel{+}. \textbf{Bolded} entries indicate distances with \uriel{+}. 
    \textbf{NA} entries indicate distances that could not be computed with \uriel{+} (see Section \ref{sec:Improve Distance Calculations}).
    }
    \label{tab:langdist-example}
\end{table}

While incorporating all of the domains mentioned above as feature categories within typological vectors (featural vectors according to \uriel{}), \uriel{} aggregates linguistic data from multiple sources to provide a standardized, all-in-one measure of language distance. 
An example of \uriel{} distances can be seen in Table \ref{tab:langdist-example}, where English and French are geographically identical, moderately distant typologically (syntactic, phonological, inventory), and genetically distant. 
This unified framework simplifies cross-linguistic comparisons by representing complex linguistic features as a single vector. By making these distances easily accessible, \uriel{} enables seamless integration into machine learning models, facilitating large-scale analysis and supporting diverse tasks, as outlined in Table \ref{tab:usage-URIEL}. 

Despite \uriel{}'s established significance in measuring language distance, several areas for improvement have been identified regarding its feature coverage and usability \citep{toossi-etal-2024-reproducibility}. For this reason, we introduce \uriel{+}, which aims to improve \uriel{}, with a focus on \emph{typological features} and \emph{typological distance}. 
Figure~\ref{fig:overview} outlines our contributions in enhancing both feature coverage and the overall usability of \uriel{}.

\begin{table*}[htp]
    \centering
    \resizebox{\textwidth}{!}{%
    \begin{tabular}{lll}
        \toprule
        \textbf{NLP Task} & \textbf{Related Papers} & \textbf{\# of citations} \\ 
        \midrule
        Cross-lingual transfer & \citealp{lin2019choosing, lauscher-etal-2020-zero, ruder-etal-2021-xtreme} & 678 \\ 
        Dependency parsing & \citealp{ustun-etal-2020-udapter} & 108 \\ 
        Machine translation & \citealp{zhang-toral-2019-effect}; \citealp{li-etal-2024-eliciting, dankers-etal-2022-transformer} & 160 \\ 
        Speech recognition  & \citealp{adams-etal-2019-massively, zanon-boito-etal-2020-mass} & 140 \\ 
        Performance prediction & \citealp{xia-etal-2020-predicting, srinivasan2021predicting, khiu-etal-2024-predicting, anugraha2024proxylm} & 65 \\
        \bottomrule
    \end{tabular}
    }
    \caption{Natural language processing (NLP) tasks, related papers utilizing \uriel{} feature vectors and/or distances, and their total forward citation counts.} 
    \label{tab:usage-URIEL}
\end{table*}


\paragraph{Feature Coverage Expansion} Currently, 31\% of the languages for which \uriel{} supports distance calculations have no data for any typological features, resulting in the use of undocumented default values \citep{toossi-etal-2024-reproducibility}. These default values prevent meaningful differentiation between languages with missing data, rendering the calculated distances unreliable \citep{khiu-etal-2024-predicting}. 
We addressed this issue by integrating additional databases into \uriel{+} (Section \ref{sec:new databases}). Incorporating these databases significantly enhances \uriel{+}'s feature coverage, increasing the number of languages available for featural distance calculations from $2724$ to $4366$.\footnote{In \uriel{}, $4005$ languages are available for distance calculation. However, only $2724$ of these languages have actual data (the remaining $1281$ rely on default values due to missing data).} Unlike the previous version, \uriel{+} now includes morphological features, which are critical for representing morphologically rich languages \citep{samardzic-etal-2024-measure}.

\paragraph{Data Integrity and Imputation} \uriel{}'s use of default values for missing data without user awareness \citep{toossi-etal-2024-reproducibility} results in distances that may not be meaningful, especially when the calculation involves low resource languages. Researchers have mitigated this by developing their own imputation methods \citep{jin-xiong-2022-informative} or by using \uriel{}’s $k$-nearest-neighbor-imputed feature vectors\footnote{\uriel{} calls these vectors $k$-nearest-neighbour aggregated vectors.} \citep{ustun-etal-2020-udapter, glavas-vulic-2021-climbing, choenni-etal-2023-cross}, although the details and quality of \uriel{}’s imputation remain undocumented. To address this issue, we integrated three well-evaluated imputation algorithms (Section \ref{sec:imputation}) and provided users with the ability to choose their preferred imputation method. This functionality  
allows distance calculations for all languages in the knowledge base and enables users to select the most suitable method for their research needs.

\paragraph{Robust Distance Calculations with Confidence Scores} 

Three issues affect \langtovec{} distance computations. First, distances are pre-computed, preventing modifications to feature vectors for updated distances. Second, \citealp{toossi-etal-2024-reproducibility} identified reproducibility issues, citing conflicting documentation on aggregation methods\footnote{In \langtovec{}, distance is computed on data that aggregates feature information for each language (using union or average aggregation). Union aggregation sets each feature value to the max value across all sources, while average aggregation sets it to the average value.} and distance metrics. Third, \langtovec{} allows distance computations for languages without values known for both languages (shared data), leading to potentially meaningless results. Furthermore, calculations use all features across all \uriel{} data sources, limiting users to one-size-fits-all calculations. To meet specific needs, researchers have manually calculated distances between subsets \citep{papadimitriou-jurafsky-2020-learning}, concatenations \citep{adams-etal-2019-massively}, or both \citep{zhang-toral-2019-effect, hossain-etal-2020-non} of \uriel{} vectors. \uriel{+} addresses these issues with a rigorous dynamic calculation system (Section \ref{sec:Improve Distance Calculations}), that allows customization of aggregation methods, metrics, features, and sources. \uriel{}'s language distances often differ from linguistic measures. For example, \uriel{} shows Croatian and Serbian as distant languages (Table \ref{tab:langdist-example}), although they should be similar \citep{samardzic-etal-2024-measure}. This mismatch could be indicative of poor data quality. Therefore, we introduce new confidence scores to assess the reliability of the distances between languages (Section \ref{sec:confidence score}).

To assess the improvement brought by the \uriel{+} enhancements, we replicated three notable downstream usages of the original \uriel{} knowledge base (see Section \ref{sec:downstream task setup}): 1) \langrank{} for selecting transfer languages in cross-lingual learning \citep{lin2019choosing}; 2) \lingualchemy{} for typological feature-driven language analysis \citep{adilazuarda2024lingualchemy}; and 3) \proxylm{} for performance prediction in multilingual settings \citep{anugraha2024proxylm}. 
Our experimental results demonstrate that \uriel{+} not only augments feature coverage and usability but also leads to better performance in practical NLP applications.

To assess how well \uriel{+} distances align with linguistic distance metrics, we conducted a case study comparing typological distances between Central and South American Indigenous languages in \uriel{} and \uriel{+} to a metric from \citealp{hammarstrom-oconnor-2013-dependency}, which accounts for predictable traits, quirks, and historical dependencies (see Section \ref{sec:casestudy}). \uriel{+} shows a higher correlation with the linguistic distances than \uriel{}, suggesting that \uriel{+} is more aligned with linguistic measurements of typological distance.

\section{From URIEL to URIEL+}

This section outlines the expansions to \uriel{} and \langtovec{}. \uriel{+} incorporates new databases to enhance feature coverage and implements imputation algorithms to handle missing values. Furthermore, \langtovec{} improves distance computations between languages and introduces new confidence scores to evaluate the reliability of these distances.

\subsection{Integrating New Databases}
\label{sec:new databases}

\begin{table*}[htp]
    \centering
    \begin{tabular}{p{2cm}p{3.5cm}p{8.5cm}}
        \toprule
        \textbf{Database} & \textbf{Reference} & \textbf{Contribution} \\ 
        \midrule
        SAPhon & \citealp{Michael_2013} & Phonology and inventories for South American indigenous languages. \\ 
        \textbf{BDPROTO} & \citealp{marsico-etal-2018-bdproto} & Phonology and inventories for ancient and reconstructed languages.\\ 
        \textbf{Grambank} & \citealp{skirgaard2023grambank} & Syntax and morphological data for $\sim$2500 languages.\\ 
        \textbf{APiCS} & \citealp{Mufwene_2013} & Typological data for pidgin and creole languages.\\ 
        \textbf{eWAVE} & \citealp{eWAVE} & Syntax and morphological data for English dialects. \\ 
        \bottomrule
    \end{tabular}
    \caption{Summary of database contributions and updates. \textbf{Bolded} entries highlight databases that have been newly added to \uriel{+} and were not present in the original \uriel{}.} 
    \label{tab:new-databases}
\end{table*}


\uriel{+} includes five additional databases (Table \ref{tab:new-databases}), incorporating data for $2898$ languages ($2858$ of which are low resource). Users can select databases (e.g., BDPROTO for ancient and reconstructed languages) based on their needs. By default, all five databases are included, providing data for $8071$ languages, with $4366$ capable of \emph{typological distance} calculations. 
Data from the five databases was first preprocessed separately, then integrated altogether into \uriel{} to create \uriel{+}. We detail the process below.

\paragraph{\textbf{Binarization for Non-Binary Features}} 
While \uriel{} (and thus, \methodname{}) supports only binary features, Grambank, APiCS, and eWAVE contain non-binary features, specifically, \emph{nominal} and \emph{ordinal variables}. \emph{Nominal variables} are categorical variables with no inherent order, while \emph{ordinal variables} represent different levels of a feature's presence in a language. To collect data from these databases, we first binarized the features. Nominal features were binarized using one-hot encoding, and ordinal features were binarized by creating a replacement feature that indicates whether that feature is present in the language. 

\paragraph{\textbf{Combining Redundant Features}} 
The Grambank, APiCS, and eWAVE databases contain features that overlap with those in \uriel{} or with each other, which can cause redundancy.

Suppose the values for a feature (say $f_1$) from one database can be inferred from another feature ($f_2$) in a different database. In such cases, we update the values in $f_1$ using inferred values from $f_2$, but not necessarily the other way around. For example, if a language has the \uriel{} feature “S\_ARTICLE\_WORD\_BEFORE\_NOUN,” it necessarily has the Grambank feature “Are there prenominal articles?” Alternatively, if $f_1$ is equivalent to $f_2$, we infer values from $f_2$ and remove $f_2$ from its database. This reduces redundancy when we integrate the preprocessed datasets into \uriel{}.


\paragraph{\textbf{Classifying and Renaming Features}} 




New features are classified as either syntactic, phonological, inventory, or morphological. This is implemented in the code by adding a prefix to the feature name. The prefixes are S\_, P\_, INV\_, and M\_ for syntactic, phonological, inventory, and morphological features, respectively - matching the conventions in the original \uriel{} knowledge base. In addition to adding these prefixes, we rename the feature names to align with the naming conventions in the original \uriel{} knowledge base. This involves capitalizing features, truncating their names, and replacing spaces with underscores.

\paragraph{\textbf{Incorporating Glottocode Identifiers}} 

\uriel{} uses ISO 639-3 codes to identify languages. While these codes remain compatible with the updated SAPhon database,\footnote{The SAPhon database identifies languages with ISO 639-3 codes.} the other databases now require glottocode identifiers \cite{forkel2022glottocodes}. Therefore, \uriel{+} employs glottocodes to better support low resource languages, including those not covered by ISO 639-3, such as Eskimo Pidgin and Singlish. Outdated ISO 639-3 codes that coexisted with their replacements in \uriel{}, such as \textit{gre} for Greek (now \textit{ell}) and \textit{alb} for Albanian (now \textit{sqi}), have been removed to ensure up-to-date and unique language identifiers.

\paragraph{Summary of Implementation Details}
\uriel{} is structured as a three-dimensional matrix, with languages, features, and data sources as the three dimensions. To incorporate data from the new databases, we extend \uriel{} by adding new entries for corresponding languages, features, and sources. Initially, these new entries are marked as missing and are subsequently updated with the available data, ensuring that \uriel{+} integrates new information while preserving existing data. For the updated SAPhon database, the new data replaces the missing values in the existing SAPhon data rather than creating any new feature columns.

\subsection{Automatic Imputation Algorithms}\label{sec:imputation}

Despite expanding \uriel{} and increasing its feature coverage, after combining the five databases into a single source, $87\%$ of values remain missing. To address this, \uriel{+} includes methods for imputing missing feature data, enabling comprehensive distance calculations between languages even with incomplete data.

We provide several imputation algorithms in \uriel{+}, including $k$-NN imputation, which was also an option in \uriel{}. Additionally, we include MIDASpy \cite{lall2023efficient}, a multiple imputation method implemented with deionizing autoencoders, and SoftImpute \cite{mazumder2010spectral}, which fits a low-rank matrix approximation via nuclear-norm regularization. 

\subsection{Robust Distance Calculations}
\label{sec:Improve Distance Calculations}

We replaced pre-computed queries with a new function that \emph{dynamically} computes distances based on current data in the knowledge base, allowing it to reflect database updates.

Since \citealp{toossi-etal-2024-reproducibility} found that union aggregation with angular distance produces distances most aligned with \uriel{}'s, these are set as the default options for computing distances. 
To resolve the documentation ambiguity, we offer users the option to choose between union or average aggregation and between cosine or angular distances, enhancing flexibility and clarity. In addition, to address the issue of meaningless distances, we exclude languages without shared data from computations. Instead, we provide imputation algorithms for users who need these distances.

We offer a distance function that computes distance based on provided features rather than all available features under a provided feature category, allowing for any combination of features. Furthermore, users can specify a particular source to use data from rather than using aggregated data. For example, one could calculate the distance of languages using $49$ syntactic features exclusively from the WALS source \citep{wals}, as was done manually in \citealp{papadimitriou-jurafsky-2020-learning}.


These changes preserve all existing functionalities while introducing feature and source customizations in calculations.




\subsection{Confidence Scores of Distance Calculations}
\label{sec:confidence score}

To evaluate the quality of the calculated distances, confidence scores are often used, as suggested by prior studies~\cite{Salati2016, Bayram2023, Bayram2024}. These scores typically aggregate several key metrics, including: 1) the amount of inaccurate data (accuracy), 2) the proportion of missing data (completeness), 3) the agreement across different sources or adherence to established constraints (consistency), 4) the recency of the data (timeliness), and 5) the deviation from a reference distribution (skewness)~\cite{Bayram2023, Batini2009}.



However, not all of these metrics are applicable or necessary in our case. Accuracy cannot be evaluated due to the absence of ground truth, timeliness is irrelevant since the data is non-temporal, and there is no natural reference distribution to measure skewness. Therefore, following \citealp{Salati2016}, our confidence scores focus solely on data completeness, data consistency, and a new metric we introduce: imputation quality. 



Formally, given languages $L_1$ and $L_2$, we define data completeness $\mathcal{M}(L_1, L_2)$ as:
\begin{equation*}
    \mathcal{M}(L_1, L_2) = 1 - \frac{p(L_1) + p(L_2)}{2}
\end{equation*}
where $p(L_i)$ is the proportion of missing values for language $L_i$.

Next, recall there may be multiple sources of feature value $j$ for language $L_i$. Let $\mathcal{S}_{i,j}$ denote the set of sources that provide values for feature $j$ on language $L_i$ and let $n_{i, j}$ be the cardinality of $\mathcal{S}_{i,j}$. Let $v_{i,j,s}$ denote the value of feature $j$ for language $L_i$ from source $s \in \mathcal{S}_{i,j}$. Define $m_{i,j}$ as the mode of the set $\{ v_{i,j,s} : s \in \mathcal{S}_{i,j} \}$. We can then define $z_{i,j}$, the number of sources that agree with the mode $m_{i,j}$, as:

\begin{equation*}
    z_{i,j} = \sum_{s \in \mathcal{S}_{i,j}} \indicator\{ v_{i,j,s} = m_{i,j} \}
\end{equation*}

where $\indicator\{\cdot\}$ is the indicator function.

Following this definition, we now define data consistency $\mathcal{C}(L_1, L_2)$ as:
\begin{equation*}
    \mathcal{C}(L_1, L_2) = \frac{a(L_1) + a(L_2)}{2}
\end{equation*}
where $a(L_i)$ is computed as:
\begin{equation*}
    a(L_i) = \frac{1}{k_i} \sum_{j=1}^{k_i} \frac{z_{i, j}}{n_{i, j}}
\end{equation*}
with $k_i$ representing the number of non-empty language features in \uriel{+} for language $L_i$. If $k_i = 0$, we set $a(L_i) = 1$.

Finally, we define imputation quality $\mathcal{I}$ as a constant $\gamma$, where $\gamma \in [0, 1]$ depends on the metric chosen to define imputation quality. For specific numerical values of $\gamma$, see Table \ref{tab:imputation-results-combined}.

\section{Validating the Knowledge Base}
\label{Experimental Setup}

Two of the ways we validated \uriel{+} were through an analysis of feature coverage and imputation quality tests on the three algorithms.

\subsection{Feature Coverage Analysis}
\label{sec:feature-coverage-improvement}

\paragraph{Experimental Setup}
To compare feature coverage between \uriel{} and \uriel{+}, we calculated the number of languages in each that have available syntactic, phonological, inventory, and morphological data. For \uriel{+}, all five databases were integrated in this comparison. Additionally, we analyzed feature coverage by categorizing languages into high resource languages (HRLs), medium resource languages (MRLs), and low resource languages (LRLs), as defined by \citealp{joshi-etal-2020-state}.

\begin{figure}[!t]
  \includegraphics[width=1.0\columnwidth]{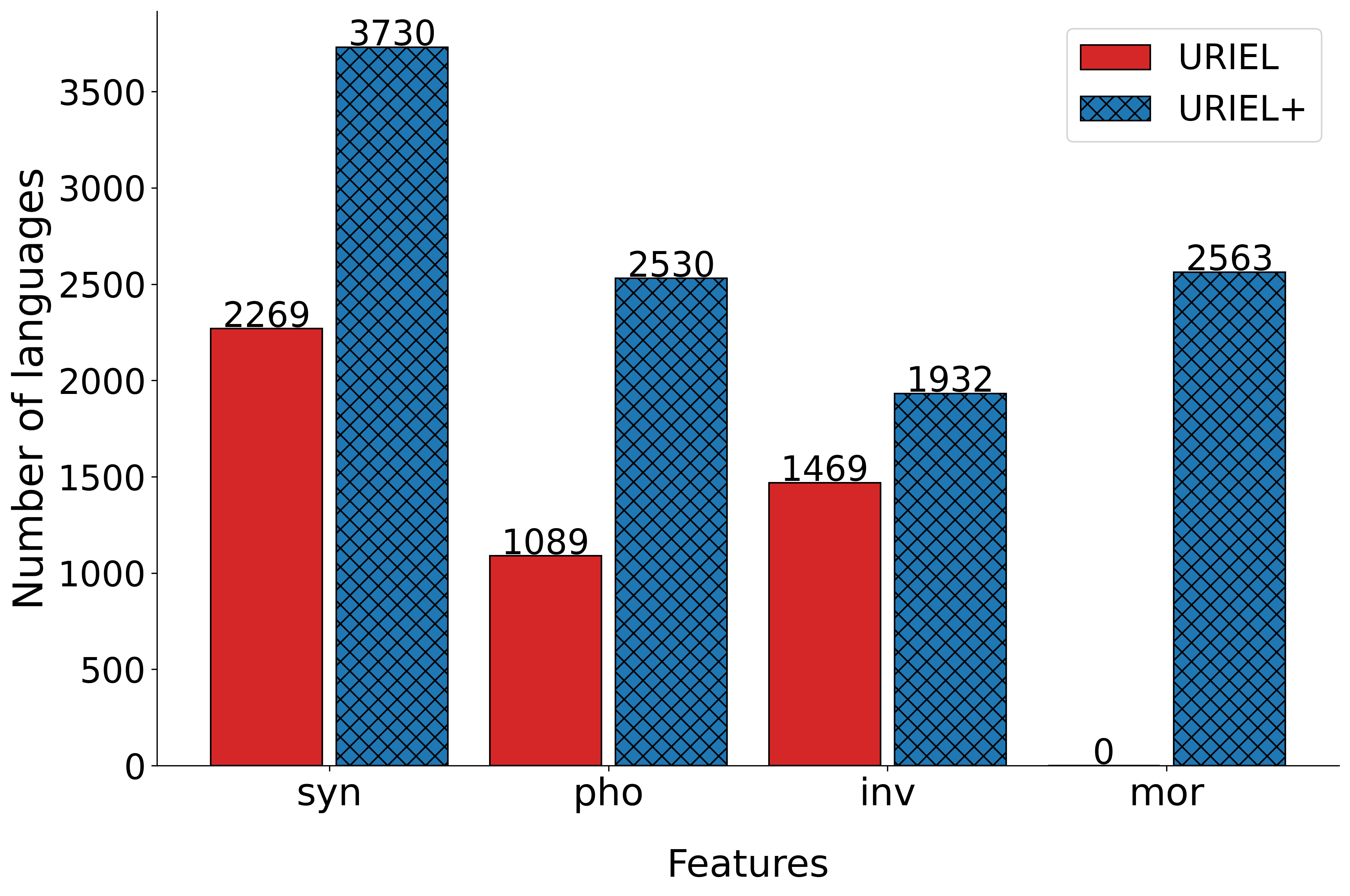}
  \caption{Number of languages with available syntactic (syn), phonological (pho), inventory (inv), and morphological (mor) data in \uriel{} and \methodname{} with all five databases.}
  \label{fig:feature_coverage}
\end{figure}

\paragraph{Results}
By integrating all five \methodname{} databases, the number of languages available for typological distance calculations increases from $4005$ to $4366$, representing a $9.01\%$  increase. This expansion includes ancient, reconstructed, pidgin, creole, and dialectal languages. Figure \ref{fig:feature_coverage} provides a detailed view of this expansion, highlighting that the number of languages with syntactic data increases from $2269$ to $3730$, a $64.39\%$ increase, driven by the inclusion of Grambank, APiCS, and eWAVE. Phonological data coverage rises from $1089$ to $2530$, representing a $132.3\%$ increase, facilitated by SAPhon, BDPROTO, APiCS. Inventory data experiences the smallest increase, expanding from $1469$ to $1932$, a $31.5\%$ increase, with contributions from SAPhon, BDPROTO, and APiCS, which mainly augment existing data.

\begin{figure*}[ht]
  \includegraphics[width=1.0\textwidth]{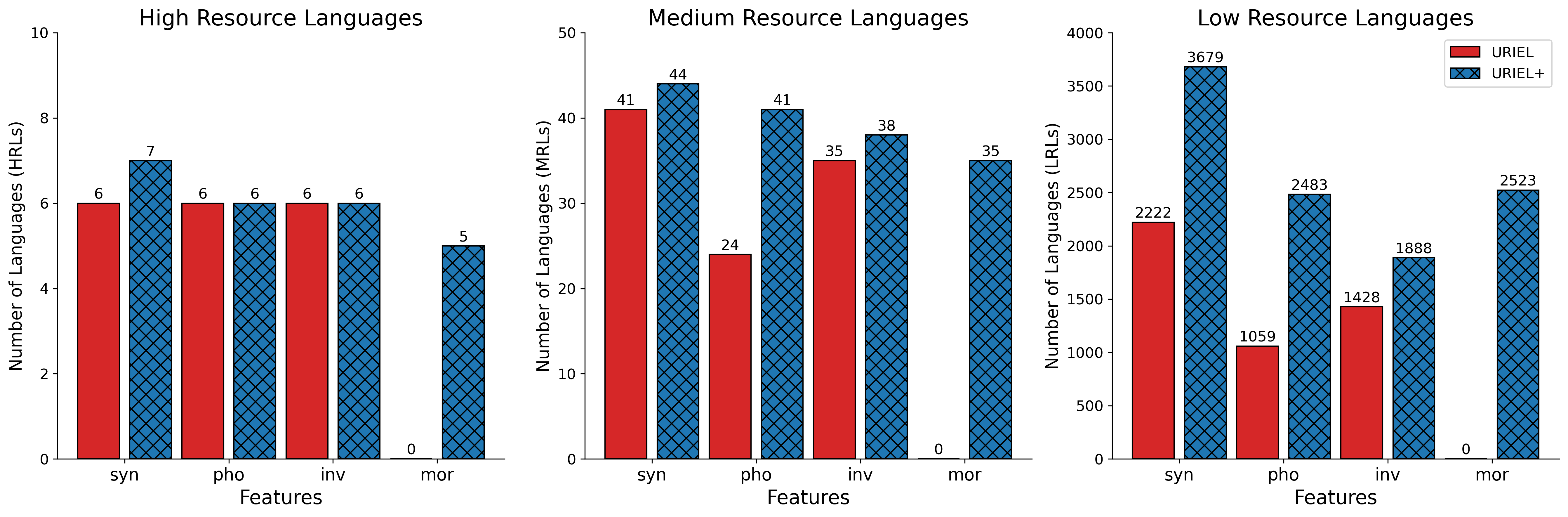}
   \caption{Number of languages with available syntactic (syn), phonological (pho), inventory (inv), and morphological (mor) data  in \uriel{} and \methodname{} with all five databases, is shown for high resource languages (HRLs), medium resource languages (MRLs), and low resource languages (LRLs) \citep{joshi-etal-2020-state} from left to right.}
    \label{fig:feature_coverage_resource_level}
\end{figure*}

Figure \ref{fig:feature_coverage_resource_level} shows the breakdown of feature coverage by language resource level. For HRLs, \uriel{+} now includes syntactic data for Arabic. The feature coverage for MRLs and LRLs improve substantially across all feature categories, with the most significant gains in phonology ($70.8\%$ more languages for MRLs and $134.5\%$ more languages for LRLs). These increases are expected, given that SAPhon, BDPROTO, and APiCS focus on LRLs and contribute substantial new phonological data. In addition, LRLs see a large increase in syntax, with a $65.6\%$ increase in the number of languages with syntactic information compared to \uriel{}. With these feature coverage expansions, \uriel{+} provides more comprehensive distance calculations to more languages. 




\subsection{Imputation Quality Test}
\paragraph{Experimental Setup}
To evaluate the three imputation algorithms and validate our choice of imputation algorithm for downstream tasks, we used the imputation quality test from \citealp{li2024comparison}. This test involves removing $20\%$ of non-missing data, imputing it, and comparing the predictions to known values using metrics like F1 for binary data and root mean square error (RMSE) for continuous data. Since the \uriel{} paper \cite{littell-etal-2017-uriel} did not include detailed metrics and procedures for its $k$-NN aggregation, the imputation quality test focuses solely on \uriel{+} and does not compare it with \uriel{}. Imputation was performed on aggregated data (union or average), with missing dialect data filled using the parent language’s data, which typically has similar typological features. 
More details on the data used and the methodology of the quality test can be found in Appendix \ref{sec:appendix-imputation}.

\paragraph{Results}

\begin{table}[!t]
    \centering
    \resizebox{.48\textwidth}{!}{
    \begin{tabular}{lcccccc}
        \toprule
        \multirow{2}{*}{\textbf{Method}} & \multicolumn{4}{c}{\textbf{Union-Agg}} & \multicolumn{2}{c}{\textbf{Average-Agg}} \\
        \cmidrule(lr){2-5} \cmidrule(lr){6-7}
         & \textbf{Accuracy} & \textbf{Precision} & \textbf{Recall} & \textbf{F1} & \textbf{RMSE} & \textbf{MAE} \\
        \midrule
        Mean & 0.8024 & 0.7248 & 0.5656 & 0.6354 & 0.3597 & 0.2608 \\
        MIDASpy & 0.8435 & 0.7819 & 0.6737 & 0.7238 & 0.3302 & 0.2171 \\
        $k$-NN & 0.8678 & 0.8136 & \textbf{0.7338} & 0.7717 & 0.3069 & \textbf{0.1809} \\
        SoftImpute & \textbf{0.8875} & \textbf{0.8801} & 0.7300 & \textbf{0.7980} & \textbf{0.2883} & 0.1886 \\
        \bottomrule
    \end{tabular}
    }
    \caption{Summary of imputation quality test results, with metrics grouped by union-aggregated and average-aggregated data. We maximise F1 on union-aggregated data and minimise RMSE on average-aggregated data. For $k$-NN, we choose $k=9$ for union-aggregated data and $k=15$ for average-aggregated data. \textbf{Bolded} entries indicate the best results in each category.}
    \label{tab:imputation-results-combined}
\end{table}
For both imputed union and average data, all imputation algorithms outperform the mean imputation as our baseline across all metrics (Table~\ref{tab:imputation-results-combined}). Comparing the performance on union-aggregated data, all algorithms have at least $6\%$ higher precision compared to the baseline. However, for each algorithm, recall is always worse than precision, meaning that each algorithm is good at imputing binary values of 1 but is too conservative in doing so. The algorithm that performs the best at maximizing both precision and recall is SoftImpute, as it has the highest F1 score. 

On average-aggregated data, all algorithms have lower RMSEs and MAEs than the baseline. However, SoftImpute is the most preferable of these algorithms. This is because SoftImpute performs almost the same as $k$-NN at minimizing errors overall since its MAE exceeds $k$-NN's by only 0.0057. However, it does much better at minimizing \textit{large} errors (it has the lowest RMSE).  

Since SoftImpute performed the best, we used \uriel{+} with union source aggregation and this method in both downstream tasks experiments (except for \lingualchemy{} where we used union- and average-aggregation) and the linguistic case study. A more granular analysis of how these imputation methods fare on specific feature categories can be found in Appendix \ref{sec:detailed-results}.

\section{URIEL+ on Downstream Tasks}

As shown in Table \ref{tab:usage-URIEL}, \uriel{} has been used in various downstream NLP tasks. Building on this, we apply \uriel{+} to different NLP tasks by comparing its performance with \uriel{} distances and vectors across three frameworks (\langrank{}, \lingualchemy{}, \proxylm{}). These frameworks are employed to evaluate multiple NLP tasks.

\subsection{Experimental Setup}
\label{sec:downstream task setup}

\paragraph{\langrank{} \cite{lin2019choosing}} \langrank{} predicts cross-lingual transfer languages using multiple data-related features, including features from all six \uriel{} distance categories. \langrank{} is evaluated on part-of-speech tagging (POS), machine translation (MT), dependency parsing (DEP), and entity linking (EL) using top-3 Normalized Discounted Cumulative Gain (NDCG@3), and shows higher average scores than other baselines.

\paragraph{\lingualchemy{} \cite{adilazuarda2024lingualchemy}}

\lingualchemy{} employs a regularization technique that utilizes \uriel{}'s syntactic and geographic vectors to guide language representations in pre-trained models. The evaluation was conducted on three tasks: semantic relatedness using SemRel2024 \cite{ousidhoum2024semrel2024collectionsemantictextual}, news classification using MasakhaNews \cite{adelani-etal-2023-masakhanews}, and intent classification using MASSIVE \cite{fitzgerald-etal-2023-massive}. Semantic relatedness was assessed using Pearson correlation, while intent and news classification were measured by accuracy. \lingualchemy{} improves performance on mBERT \cite{devlin2019bertpretrainingdeepbidirectional} and XLM-R \cite{conneau2020unsupervisedcrosslingualrepresentationlearning}, benefiting LRLs and unseen language generalization.

\begin{table}[!htbp]
\centering
\resizebox{.49\textwidth}{!}{
    \begin{tabular}{lrl@{\hspace{0.5em}}l@{ }r@{ }l}
    \midrule
    \textbf{Downstream Tasks} & \textbf{Subtasks} & \textbf{\uriel{}} & \textbf{\methodname{}} \\
    \midrule
    \multirow{4}{*}{\langrank{}} & POS & 22.9 & 24.6 (\textcolor{darkblue}{$\uparrow 7.42\%$}) \\ 
    & MT & 42.55 & \textbf{45.45} (\textcolor{darkblue}{$\uparrow 6.82\%$}) \\
    & DEP & 71.7 & 72.85 (\textcolor{darkblue}{$\uparrow 1.60\%$}) \\ 
    & EL & 66.50 & 66.80 (\textcolor{darkblue}{$\uparrow 0.45\%$}) \\
    \midrule
    \multirow{3}{*}{\lingualchemy{}} & SemRel2024 & 0.008 & \textbf{0.012} (\textcolor{darkblue}{$\uparrow 50.00\%$}) \\
    & MasakhaNews & 76.13 & \textbf{76.43} (\textcolor{darkblue}{$\uparrow 0.39\%$}) \\
    & MASSIVE & 72.95 & \textbf{73.16} (\textcolor{darkblue} {$\uparrow 0.29\%$}) \\ \midrule
    \multirow{3}{*}{\proxylm{}}  & Unseen & 4.35 & \textbf{4.08} (\textcolor{darkblue}{$\uparrow 6.21\%$}) \\
    & Random & 2.92 & 2.82 (\textcolor{darkblue}{$\uparrow 3.42\%$})  \\
    & LOLO & 3.84 & 3.85 (\textcolor{darkred}{$\downarrow 0.26\%$}) \\
   
    \bottomrule
    \end{tabular}
}
\caption{
Results for each subtasks in \langrank{}, \lingualchemy{}, and \proxylm{}. \textbf{\uriel{}} refers to tasks using the \uriel{} knowledge base, and \textbf{\methodname{}} to tasks using the \methodname{} knowledge base. Higher values indicate better performance for \langrank{} and \lingualchemy{}, while lower values do so for \proxylm{}. \textcolor{darkblue}{$\uparrow$} indicates an improvement, and \textcolor{darkred}{$\downarrow$} signifies a decline relative to \uriel{}. \textbf{Bolded} numbers indicate results that are statistically significant at a 0.05 significance level using the Wilcoxon signed-rank test \cite{Rey2014}. More detail about the metrics used along with the breakdown of the results can be found in Appendices~\ref{sec:appendixE} and ~\ref{sec:appendix-performance}, respectively.}
\label{tab:results-summary}
\end{table}

\paragraph{\proxylm{} \cite{anugraha2024proxylm}} 

\proxylm{} estimates the performance of language models in multilingual NLP tasks, using proxy models and \uriel{} distances without fine-tuning, saving time and computational resources. \proxylm{} outperforms state-of-the-art performance prediction methods in terms of RMSE, while also demonstrating robustness and efficiency. \proxylm{} was evaluated on M2M100 \cite{fan2020englishcentricmultilingualmachinetranslation} and NLLB \cite{costa2022no} under three settings: Unseen (generalization on unseen languages to the pre-trained language models), Random (random split), and LOLO (Leave-One-Language-Out).

For each of the frameworks, we will refer to each of the tasks, datasets, or settings evaluations as “subtasks” for convenience. Further details on the experimental settings for each of the frameworks along with the downstream tasks can be found in Appendix \ref{sec:appendixE}.

\subsection{Results}

Table \ref{tab:results-summary} compares the summary of \uriel{} and \methodname{} on the three downstream tasks. \langrank{} with \uriel{+} distances consistently outperforms \uriel{} distances across all subtasks. The most significant performance gain is observed in the POS subtask, followed by MT, DEP, and EL (detailed breakdown in Table \ref{tab:results-langrank1} in the Appendix), which compares \langrank{} using only language features (referred to as \langrank{} (lang feats)) and both dataset and language features (referred to as \langrank{} (all)).

Both \langrank{} (lang feats) and \langrank{} (all) show performance gains in the POS and MT subtasks, with a notable boost in MT from including dataset features. This is because MT often emphasizes data-specific characteristics over linguistic features, resulting in a larger improvement for \langrank{} with all data and language features rather than with only language features \cite{lin2019choosing}. For the EL subtask, \langrank{} (all) performs slightly worse, likely due to the absence of dataset-specific information. Interestingly, using only language features improves performance for the EL subtask. In the DEP subtask, integrating both dataset and language features enhances performance, reinforcing the value of combining \uriel{+} features with dataset-specific information \cite{lin2019choosing}.

In \lingualchemy{}, $\methodname{}$ shows a slight performance increase across all subtasks, highlighting its advantage. Table \ref{tab:results-lingualchemy} in the Appendix indicates that average syntax vectors yield mixed results, with both increases and decreases in performance compared to syntax $k$-NN vectors. This suggests that syntax $k$-NN vectors, which used union-aggregation, retain more feature diversity by including any available presence. Meanwhile, average aggregation might lose key distinctions since it smooths out differences between features. Therefore, $k$-NN vectors are more reliable when incorporating $\methodname{}$ features for \lingualchemy{}.

For \proxylm{}, competitive performance is observed, with significant improvements in Unseen and Random settings, as detailed in Tables \ref{tab:results-summary} and \ref{tab:results-proxylm} in the Appendix. These results suggest that \uriel{+} features improve generalization to previously unseen languages. The insignificant change in performance for the LOLO setting may indicate that the existing \uriel{} features across multiple languages are already sufficient for maintaining robust predictions when leaving one language out from the regressor.

The Wilcoxon signed-rank test \cite{Rey2014} indicates that most results in Table \ref{tab:results-summary} are statistically significant ($p$-values less than 0.05), except for \proxylm{}’s Random and LOLO settings, and \langrank{}’s POS, EL, and DEP tasks. The non-significance of \proxylm{}’s LOLO setting suggests that the small performance drop is negligible. In contrast, the non-significance of \langrank{}’s POS, EL, and DEP tasks is likely due to the limited number of target languages in \langrank{}.

Overall, $\methodname{}$ demonstrates competitive performance across all tasks compared to \uriel{}, especially where language-specific features are crucial.

\begin{table}[!ht]
    \centering
    \resizebox{.49\textwidth}{!}{%
    \begin{tabular}{lccc}
    \toprule
    \textbf{Language Pair} & \textbf{URIEL} & \textbf{\uriel{+}} & $\mathbf{G_d}$ \\
    \midrule
    Paez-Bintucua & 0.50 & 0.55 & 0.49 \\
    Sambu-Cayapa & 0.50 & 0.54 & 0.48 \\
    Ulua-Paez & 0.70 & 0.74 & 0.45 \\
    Sumo-Paez & 0.70 & 0.74 & 0.45 \\
    Paez-Misquito & 0.80 & 0.58 & 0.43 \\
    Quiche-Paez & 0.60 & 0.64 & 0.43 \\
    Quiche-Boruca & 0.80 & 0.65 & 0.41 \\
    Quiche-Huaunana & 0.60 & 0.61 & 0.40 \\
    Xinca-Cofan & 0.70 & 0.73 & 0.39 \\
    Xinca-Boruca & 0.90 & 0.79 & 0.39 \\
    Quiche-Colorado & 0.60 & 0.64 & 0.38 \\
    Quiche-Cayapa & 0.60 & 0.62 & 0.36 \\
    Quiche-Cuna & 0.80 & 0.56 & 0.35 \\
    Paya-Bintucua & 0.60 & 0.50 & 0.35 \\
    Cuna-Boruca & 0.00 & 0.51 & 0.35 \\
    Paya-Muisca & 0.60 & 0.51 & 0.33 \\
    Huaunana-Boruca & 0.80 & 0.51 & 0.32 \\
    Paya-Cagaba & 0.50 & 0.43 & 0.31 \\
    Xinca-Camsa & 0.70 & 0.76 & 0.30 \\
    Quiche-Lenca & 0.70 & 0.73 & 0.28 \\
    \midrule
    \textbf{Rank Correlation with $\mathbf{G_d}$} & $-0.05$ & \textbf{0.19} & N/A \\
    \bottomrule
    \end{tabular}%
    }
    \caption{Language distances from \uriel{}, \uriel{+} and the dependency-sensitive Gower coefficient ($G_d$), and rank correlation of \uriel{} and \uriel{+} with $G_d$.}
    \label{tab:results-casestudy}
\end{table}

\section{Distance Alignment Case Study}

In this section, we demonstrate that the distances provided by \uriel{+} align more closely with a linguistic distance metric. This is illustrated through a case study on distance measures between two languages, as conducted by \citealp{hammarstrom-oconnor-2013-dependency}.

\subsection{Experimental Setup}

\label{sec:casestudy}

We assessed the accuracy of \uriel{} and \uriel{+} distance measures by comparing them to a modified Gower coefficient, $G_d$ \citep{56fe9d33-d905-3182-860c-ed85b7b00402}, for typological distance \citep{hammarstrom-oconnor-2013-dependency}. This metric weights scores based on both predictable and idiosyncratic traits, while accounting for dependencies and historical contact. The case study focused on Isthmo-Colombian languages, which now have more data in \uriel{+}, due to updates in the SAPhon database. We evaluated which knowledge base aligns better with $G_d$ by using Kendall's rank correlation. This approach was chosen over direct comparisons with $G_d$ distances due to differences in unit scales, opting instead for unit-agnostic rank correlation. To compare distances from \uriel{} and \uriel{+} with those derived from the coefficient, we computed featural distances using \langtovec{}.\footnote{The ISO 639-3 codes and glottocodes required for these distances are provided in the Appendix, Table \ref{tab:langs-casestudy}}


\subsection{Results}

Table \ref{tab:results-casestudy} shows the distances for each language pair in the case study, where Kendall's rank correlations between $G_d$ and the distances from \uriel{} and \uriel{+} are $-0.05$ and $0.19$ respectively. 

We used the Perm-Both hypothesis test \cite{deutsch2021statistical} to compare the significance between these two correlations. The $p$-value was found to be $0.307$ which is not statistically significant at the $0.05$ level. The lack of significance is likely due to the small sample size, as smaller datasets tend to have higher $p$-values \cite{johnson1999insignificance}. Comparing additional language pairs could help further validate these findings. Although the difference in correlation is not statistically significant, \uriel{+} shows a trend toward better alignment with $G_d$, suggesting that it may improve the alignment of typological distance metrics in NLP with those used in linguistics.

\section{Conclusion}
We introduce \uriel{+}, an enhanced knowledge base that expands the coverage of typological features by integrating five additional databases, providing data for $2858$ LRLs. Furthermore, \uriel{+} improves the robustness and usability of \langtovec{} distances through carefully selected imputation methods, a rigorous study of appropriate distance calculations, and the establishment of new confidence scores to validate distance reliability. In addition, we demonstrate \uriel{+}'s competitive performance on downstream NLP tasks and its closer alignment with real-world linguistic distances through a case study. These improvements are critical for a wide range of multilingual applications and contribute to the linguistic inclusion of LRLs. As an open-source tool, we hope the community will contribute to its ongoing improvements and database expansions in the future.








\section*{Limitations}

With the new features from Grambank, APiCS, and eWAVE databases, \uriel{+} predominantly emphasizes syntactic ($474$ features) and morphological ($133$ features) data, with fewer contributions to phonological ($30$ features) and inventory ($163$ features) data. This shift skews the focus of featural distance towards grammar ($607$ features overall) over sound, potentially underrepresenting phonological aspects ($193$ features overall). To address this imbalance, we plan to include more phonological features, creating more specific distinctions based on current features in future work. 



Similar to \uriel, \uriel{+} does not have information on language scripts. To address this, we will introduce scripts as a feature category using ScriptSource (\citealp{Holloway-2012}), which covers $8290$ languages in future work. 

Other future work would be integrating \uriel{+} as an external knowledge base with large language models, merging structured knowledge with flexible language modeling.

\section*{Acknowledgement}
We thank Hasti Toossi and Guo Qing Huai for their valuable feedback. We also thank Dr. Patrick Littell (NRCC/Canada) for his insightful discussion and suggestions. 


\bibliography{references}

\begin{thebibliography}{59}
\providecommand{\natexlab}[1]{#1}

\bibitem[{Adams et~al.(2019)Adams, Wiesner, Watanabe, and Yarowsky}]{adams-etal-2019-massively}
Oliver Adams, Matthew Wiesner, Shinji Watanabe, and David Yarowsky. 2019.
\newblock \href {https://doi.org/10.18653/v1/N19-1009} {Massively multilingual adversarial speech recognition}.
\newblock In \emph{Proceedings of the 2019 Conference of the North {A}merican Chapter of the Association for Computational Linguistics: Human Language Technologies, Volume 1 (Long and Short Papers)}, pages 96--108, Minneapolis, Minnesota. Association for Computational Linguistics.

\bibitem[{Adelani et~al.(2023)Adelani, Masiak, Azime, Alabi, Tonja, Mwase, Ogundepo, Dossou, Oladipo, Nixdorf, Emezue, Al-azzawi, Sibanda, David, Ndolela, Mukiibi, Ajayi, Moteu, Odhiambo, Owodunni, Obiefuna, Mohamed, Muhammad, Ababu, Salahudeen, Yigezu, Gwadabe, Abdulmumin, Taye, Awoyomi, Shode, Adelani, Abdulganiyu, Omotayo, Adeeko, Afolabi, Aremu, Samuel, Siro, Kimotho, Ogbu, Mbonu, Chukwuneke, Fanijo, Ojo, Awosan, Kebede, Sakayo, Nyatsine, Sidume, Yousuf, Oduwole, Tshinu, Kimanuka, Diko, Nxakama, Nigusse, Johar, Mohamed, Hassan, Mehamed, Ngabire, Jules, Ssenkungu, and Stenetorp}]{adelani-etal-2023-masakhanews}
David~Ifeoluwa Adelani, Marek Masiak, Israel~Abebe Azime, Jesujoba Alabi, Atnafu~Lambebo Tonja, Christine Mwase, Odunayo Ogundepo, Bonaventure F.~P. Dossou, Akintunde Oladipo, Doreen Nixdorf, Chris~Chinenye Emezue, Sana Al-azzawi, Blessing Sibanda, Davis David, Lolwethu Ndolela, Jonathan Mukiibi, Tunde Ajayi, Tatiana Moteu, Brian Odhiambo, Abraham Owodunni, Nnaemeka Obiefuna, Muhidin Mohamed, Shamsuddeen~Hassan Muhammad, Teshome~Mulugeta Ababu, Saheed~Abdullahi Salahudeen, Mesay~Gemeda Yigezu, Tajuddeen Gwadabe, Idris Abdulmumin, Mahlet Taye, Oluwabusayo Awoyomi, Iyanuoluwa Shode, Tolulope Adelani, Habiba Abdulganiyu, Abdul-Hakeem Omotayo, Adetola Adeeko, Abeeb Afolabi, Anuoluwapo Aremu, Olanrewaju Samuel, Clemencia Siro, Wangari Kimotho, Onyekachi Ogbu, Chinedu Mbonu, Chiamaka Chukwuneke, Samuel Fanijo, Jessica Ojo, Oyinkansola Awosan, Tadesse Kebede, Toadoum~Sari Sakayo, Pamela Nyatsine, Freedmore Sidume, Oreen Yousuf, Mardiyyah Oduwole, Kanda Tshinu, Ussen Kimanuka, Thina Diko, Siyanda Nxakama, Sinodos
  Nigusse, Abdulmejid Johar, Shafie Mohamed, Fuad~Mire Hassan, Moges~Ahmed Mehamed, Evrard Ngabire, Jules Jules, Ivan Ssenkungu, and Pontus Stenetorp. 2023.
\newblock \href {https://doi.org/10.18653/v1/2023.ijcnlp-main.10} {{M}asakha{NEWS}: News topic classification for {A}frican languages}.
\newblock In \emph{Proceedings of the 13th International Joint Conference on Natural Language Processing and the 3rd Conference of the Asia-Pacific Chapter of the Association for Computational Linguistics (Volume 1: Long Papers)}, pages 144--159, Nusa Dua, Bali. Association for Computational Linguistics.

\bibitem[{Adilazuarda et~al.(2024)Adilazuarda, Cahyawijaya, Aji, Winata, and Purwarianti}]{adilazuarda2024lingualchemy}
Muhammad~Farid Adilazuarda, Samuel Cahyawijaya, Alham~Fikri Aji, Genta~Indra Winata, and Ayu Purwarianti. 2024.
\newblock Lingualchemy: Fusing typological and geographical elements for unseen language generalization.
\newblock \emph{arXiv preprint arXiv:2401.06034}.

\bibitem[{Anugraha et~al.(2024)Anugraha, Winata, Li, Irawan, and Lee}]{anugraha2024proxylm}
David Anugraha, Genta~Indra Winata, Chenyue Li, Patrick~Amadeus Irawan, and En-Shiun~Annie Lee. 2024.
\newblock Proxylm: Predicting language model performance on multilingual tasks via proxy models.
\newblock \emph{arXiv preprint arXiv:2406.09334}.

\bibitem[{Batini et~al.(2009)Batini, Cappiello, Francalanci, and Maurino}]{Batini2009}
Carlo Batini, Cinzia Cappiello, Chiara Francalanci, and Andrea Maurino. 2009.
\newblock \href {https://doi.org/10.1145/1541880.1541883} {Methodologies for data quality assessment and improvement}.
\newblock \emph{ACM Computing Surveys}, 41(3).

\bibitem[{Bayram et~al.(2024)Bayram, Ahmed, and Hallin}]{Bayram2024}
Firas Bayram, Bestoun~S. Ahmed, and Erik Hallin. 2024.
\newblock \href {https://doi.org/10.1016/j.jss.2024.112184} {Adaptive data quality scoring operations framework using drift-aware mechanism for industrial applications}.
\newblock \emph{Journal of Systems and Software}, 217:112184.

\bibitem[{Bayram et~al.(2023)Bayram, Ahmed, Hallin, and Engman}]{Bayram2023}
Firas Bayram, Bestoun~S. Ahmed, Erik Hallin, and Anton Engman. 2023.
\newblock \href {https://doi.org/10.1145/3593434.3593445} {{DQSOps: Data Quality Scoring Operations Framework for Data-Driven Applications}}.
\newblock In \emph{EASE '23: Proceedings of the International Conference on Evaluation and Assessment in Software Engineering}, Oulu, Finland. Association for Computing Machinery.

\bibitem[{Bradlow et~al.(2010)Bradlow, Clopper, Smiljanic, and Walter}]{BRADLOW2010930}
Ann Bradlow, Cynthia Clopper, Rajka Smiljanic, and Mary~Ann Walter. 2010.
\newblock \href {https://doi.org/10.1016/j.specom.2010.06.003} {A perceptual phonetic similarity space for languages: Evidence from five native language listener groups}.
\newblock \emph{Speech Communication}, 52(11):930--942.
\newblock Non-native Speech Perception in Adverse Conditions.

\bibitem[{Cahyawijaya et~al.(2023)Cahyawijaya, Lovenia, Koto, Adhista, Dave, Oktavianti, Akbar, Lee, Shadieq, Cenggoro, Linuwih, Wilie, Muridan, Winata, Moeljadi, Aji, Purwarianti, and Fung}]{cahyawijaya-etal-2023-nusawrites}
Samuel Cahyawijaya, Holy Lovenia, Fajri Koto, Dea Adhista, Emmanuel Dave, Sarah Oktavianti, Salsabil Akbar, Jhonson Lee, Nuur Shadieq, Tjeng~Wawan Cenggoro, Hanung Linuwih, Bryan Wilie, Galih Muridan, Genta Winata, David Moeljadi, Alham~Fikri Aji, Ayu Purwarianti, and Pascale Fung. 2023.
\newblock \href {https://doi.org/10.18653/v1/2023.ijcnlp-main.60} {{N}usa{W}rites: Constructing high-quality corpora for underrepresented and extremely low-resource languages}.
\newblock In \emph{Proceedings of the 13th International Joint Conference on Natural Language Processing and the 3rd Conference of the Asia-Pacific Chapter of the Association for Computational Linguistics (Volume 1: Long Papers)}, pages 921--945, Nusa Dua, Bali. Association for Computational Linguistics.

\bibitem[{Choenni et~al.(2023)Choenni, Garrette, and Shutova}]{choenni-etal-2023-cross}
Rochelle Choenni, Dan Garrette, and Ekaterina Shutova. 2023.
\newblock \href {https://doi.org/10.1162/coli_a_00482} {Cross-lingual transfer with language-specific subnetworks for low-resource dependency parsing}.
\newblock \emph{Computational Linguistics}, pages 613--641.

\bibitem[{Christina~Nelson and Wrembel(2021)}]{NelsonMultilingual2021}
Halina~Lewandowska Christina~Nelson, Iga~Krzysik and Magdalena Wrembel. 2021.
\newblock \href {https://doi.org/10.1080/09658416.2021.1897132} {Multilingual learners’ perceptions of cross-linguistic distances: a proposal for a visual psychotypological measure}.
\newblock \emph{Language Awareness}, 30(2):176--194.

\bibitem[{Conneau et~al.(2020)Conneau, Khandelwal, Goyal, Chaudhary, Wenzek, Guzmán, Grave, Ott, Zettlemoyer, and Stoyanov}]{conneau2020unsupervisedcrosslingualrepresentationlearning}
Alexis Conneau, Kartikay Khandelwal, Naman Goyal, Vishrav Chaudhary, Guillaume Wenzek, Francisco Guzmán, Edouard Grave, Myle Ott, Luke Zettlemoyer, and Veselin Stoyanov. 2020.
\newblock \href {https://arxiv.org/abs/1911.02116} {Unsupervised cross-lingual representation learning at scale}.
\newblock \emph{Preprint}, arXiv:1911.02116.

\bibitem[{Costa-juss{\`a} et~al.(2022)Costa-juss{\`a}, Cross, {\c{C}}elebi, Elbayad, Heafield, Heffernan, Kalbassi, Lam, Licht, Maillard et~al.}]{costa2022no}
Marta~R Costa-juss{\`a}, James Cross, Onur {\c{C}}elebi, Maha Elbayad, Kenneth Heafield, Kevin Heffernan, Elahe Kalbassi, Janice Lam, Daniel Licht, Jean Maillard, et~al. 2022.
\newblock No language left behind: Scaling human-centered machine translation.
\newblock \emph{arXiv preprint arXiv:2207.04672}.

\bibitem[{Dankers et~al.(2022)Dankers, Lucas, and Titov}]{dankers-etal-2022-transformer}
Verna Dankers, Christopher Lucas, and Ivan Titov. 2022.
\newblock \href {https://doi.org/10.18653/v1/2022.acl-long.252} {Can transformer be too compositional? analysing idiom processing in neural machine translation}.
\newblock In \emph{Proceedings of the 60th Annual Meeting of the Association for Computational Linguistics (Volume 1: Long Papers)}, pages 3608--3626, Dublin, Ireland. Association for Computational Linguistics.

\bibitem[{Deutsch et~al.(2021)Deutsch, Dror, and Roth}]{deutsch2021statistical}
Daniel Deutsch, Rotem Dror, and Dan Roth. 2021.
\newblock A statistical analysis of summarization evaluation metrics using resampling methods.
\newblock \emph{Transactions of the Association for Computational Linguistics}, 9:1132--1146.

\bibitem[{Devlin et~al.(2019)Devlin, Chang, Lee, and Toutanova}]{devlin2019bertpretrainingdeepbidirectional}
Jacob Devlin, Ming-Wei Chang, Kenton Lee, and Kristina Toutanova. 2019.
\newblock \href {https://arxiv.org/abs/1810.04805} {Bert: Pre-training of deep bidirectional transformers for language understanding}.
\newblock \emph{Preprint}, arXiv:1810.04805.

\bibitem[{Dryer and Haspelmath(2013)}]{wals}
Matthew~S. Dryer and Martin Haspelmath, editors. 2013.
\newblock \href {https://doi.org/10.5281/zenodo.7385533} {\emph{WALS Online (v2020.3)}}.
\newblock Zenodo.

\bibitem[{Fan et~al.(2020)Fan, Bhosale, Schwenk, Ma, El-Kishky, Goyal, Baines, Celebi, Wenzek, Chaudhary, Goyal, Birch, Liptchinsky, Edunov, Grave, Auli, and Joulin}]{fan2020englishcentricmultilingualmachinetranslation}
Angela Fan, Shruti Bhosale, Holger Schwenk, Zhiyi Ma, Ahmed El-Kishky, Siddharth Goyal, Mandeep Baines, Onur Celebi, Guillaume Wenzek, Vishrav Chaudhary, Naman Goyal, Tom Birch, Vitaliy Liptchinsky, Sergey Edunov, Edouard Grave, Michael Auli, and Armand Joulin. 2020.
\newblock \href {https://arxiv.org/abs/2010.11125} {Beyond english-centric multilingual machine translation}.
\newblock \emph{Preprint}, arXiv:2010.11125.

\bibitem[{FitzGerald et~al.(2023)FitzGerald, Hench, Peris, Mackie, Rottmann, Sanchez, Nash, Urbach, Kakarala, Singh, Ranganath, Crist, Britan, Leeuwis, Tur, and Natarajan}]{fitzgerald-etal-2023-massive}
Jack FitzGerald, Christopher Hench, Charith Peris, Scott Mackie, Kay Rottmann, Ana Sanchez, Aaron Nash, Liam Urbach, Vishesh Kakarala, Richa Singh, Swetha Ranganath, Laurie Crist, Misha Britan, Wouter Leeuwis, Gokhan Tur, and Prem Natarajan. 2023.
\newblock \href {https://doi.org/10.18653/v1/2023.acl-long.235} {{MASSIVE}: A 1{M}-example multilingual natural language understanding dataset with 51 typologically-diverse languages}.
\newblock In \emph{Proceedings of the 61st Annual Meeting of the Association for Computational Linguistics (Volume 1: Long Papers)}, pages 4277--4302, Toronto, Canada. Association for Computational Linguistics.

\bibitem[{Forkel et~al.(2022)}]{forkel2022glottocodes}
Robert Forkel et~al. 2022.
\newblock Glottocodes: Identifiers linking families, languages and dialects to comprehensive reference information.
\newblock \emph{Semantic Web}, 13(6):917--924.

\bibitem[{Gamallo et~al.(2017)Gamallo, Pichel, and Alegria}]{gamallo-pichel-alegria-2017-language-id}
Pablo Gamallo, José~Ramom Pichel, and Iñaki Alegria. 2017.
\newblock \href {https://doi.org/10.1016/j.physa.2017.05.011} {From language identification to language distance}.
\newblock \emph{Physica A: Statistical Mechanics and its Applications}, 484:152--162.

\bibitem[{Glava{\v{s}} and Vuli{\'c}(2021)}]{glavas-vulic-2021-climbing}
Goran Glava{\v{s}} and Ivan Vuli{\'c}. 2021.
\newblock \href {https://doi.org/10.18653/v1/2021.findings-acl.431} {Climbing the tower of treebanks: Improving low-resource dependency parsing via hierarchical source selection}.
\newblock In \emph{Findings of the Association for Computational Linguistics: ACL-IJCNLP 2021}, pages 4878--4888, Online. Association for Computational Linguistics.

\bibitem[{Gowda et~al.(2021)Gowda, Zhang, Mattmann, and May}]{gowda-etal-2021-many}
Thamme Gowda, Zhao Zhang, Chris Mattmann, and Jonathan May. 2021.
\newblock \href {https://doi.org/10.18653/v1/2021.acl-demo.37} {Many-to-{E}nglish machine translation tools, data, and pretrained models}.
\newblock In \emph{Proceedings of the 59th Annual Meeting of the Association for Computational Linguistics and the 11th International Joint Conference on Natural Language Processing: System Demonstrations}, pages 306--316, Online. Association for Computational Linguistics.

\bibitem[{Gower(1971)}]{56fe9d33-d905-3182-860c-ed85b7b00402}
J.~C. Gower. 1971.
\newblock \href {http://www.jstor.org/stable/2528823} {A general coefficient of similarity and some of its properties}.
\newblock \emph{Biometrics}, 27(4):857--871.

\bibitem[{Hammarström and O’Connor(2013)}]{hammarstrom-oconnor-2013-dependency}
Harald Hammarström and Loretta O’Connor. 2013.
\newblock \href {https://doi.org/10.1515/9783110305258.329} {\emph{Dependency-sensitive typological distance}}, pages 329--352.
\newblock Walter de Gruyter.

\bibitem[{Haspelmath(2020)}]{haspelmath-2020-structural-uniqueness}
Martin Haspelmath. 2020.
\newblock \href {https://doi.org/10.1075/alal.20032.has} {The structural uniqueness of languages and the value of comparison for language description}.
\newblock \emph{Asian Languages and Linguistics}, 1:346--366.

\bibitem[{Haspelmath(2023)}]{haspelmath-2023-word-class-universals}
Martin Haspelmath. 2023.
\newblock \href {https://doi.org/10.1093/oxfordhb/9780198852889.013.2} {\emph{Word-Class Universals and Language-Particular Analysis}}, pages 15--40.
\newblock Oxford University Press.

\bibitem[{Holloway(2013)}]{Holloway-2012}
Steph Holloway. 2013.
\newblock \href {https://scriptsource.org/cms/scripts/page.php} {Scriptsource - writing systems, computers and people}.

\bibitem[{Hossain et~al.(2020)Hossain, Anastasopoulos, Blanco, and Palmer}]{hossain-etal-2020-non}
Md~Mosharaf Hossain, Antonios Anastasopoulos, Eduardo Blanco, and Alexis Palmer. 2020.
\newblock \href {https://doi.org/10.18653/v1/2020.findings-emnlp.345} {It{'}s not a non-issue: Negation as a source of error in machine translation}.
\newblock In \emph{Findings of the Association for Computational Linguistics: EMNLP 2020}, pages 3869--3885, Online. Association for Computational Linguistics.

\bibitem[{Jin and Xiong(2022)}]{jin-xiong-2022-informative}
Renren Jin and Deyi Xiong. 2022.
\newblock \href {https://aclanthology.org/2022.coling-1.458} {Informative language representation learning for massively multilingual neural machine translation}.
\newblock In \emph{Proceedings of the 29th International Conference on Computational Linguistics}, pages 5158--5174, Gyeongju, Republic of Korea. International Committee on Computational Linguistics.

\bibitem[{Johnson(1999)}]{johnson1999insignificance}
Douglas~H. Johnson. 1999.
\newblock The insignificance of statistical significance testing.
\newblock \emph{Journal of Wildlife Management}, 63(3):763--772.

\bibitem[{Joshi et~al.(2020)Joshi, Santy, Budhiraja, Bali, and Choudhury}]{joshi-etal-2020-state}
Pratik Joshi, Sebastin Santy, Amar Budhiraja, Kalika Bali, and Monojit Choudhury. 2020.
\newblock \href {https://doi.org/10.18653/v1/2020.acl-main.560} {The state and fate of linguistic diversity and inclusion in the {NLP} world}.
\newblock In \emph{Proceedings of the 58th Annual Meeting of the Association for Computational Linguistics}, pages 6282--6293, Online. Association for Computational Linguistics.

\bibitem[{Khiu et~al.(2024)Khiu, Toossi, Liu, Li, Anugraha, Flores, Roman, Do{\u{g}}ru{\"o}z, and Lee}]{khiu-etal-2024-predicting}
Eric Khiu, Hasti Toossi, Jinyu Liu, Jiaxu Li, David Anugraha, Juan Flores, Leandro Roman, A.~Seza Do{\u{g}}ru{\"o}z, and En-Shiun Lee. 2024.
\newblock \href {https://aclanthology.org/2024.findings-eacl.100} {Predicting machine translation performance on low-resource languages: The role of domain similarity}.
\newblock In \emph{Findings of the Association for Computational Linguistics: EACL 2024}, pages 1474--1486, St. Julian{'}s, Malta. Association for Computational Linguistics.

\bibitem[{Kortmann et~al.(2020)Kortmann, Lunkenheimer, and Ehret}]{eWAVE}
Bernd Kortmann, Kerstin Lunkenheimer, and Katharina Ehret, editors. 2020.
\newblock \href {https://ewave-atlas.org/} {\emph{eWAVE}}.
\newblock Zenodo.

\bibitem[{Lall and Robinson(2023)}]{lall2023efficient}
Ranjit Lall and Thomas Robinson. 2023.
\newblock Efficient multiple imputation for diverse data in python and r: Midaspy and rmidas.
\newblock \emph{Journal of Statistical Software}, 107:1--38.

\bibitem[{Lauscher et~al.(2020)Lauscher, Ravishankar, Vuli{\'c}, and Glava{\v{s}}}]{lauscher-etal-2020-zero}
Anne Lauscher, Vinit Ravishankar, Ivan Vuli{\'c}, and Goran Glava{\v{s}}. 2020.
\newblock \href {https://doi.org/10.18653/v1/2020.emnlp-main.363} {From zero to hero: {O}n the limitations of zero-shot language transfer with multilingual {T}ransformers}.
\newblock In \emph{Proceedings of the 2020 Conference on Empirical Methods in Natural Language Processing (EMNLP)}, pages 4483--4499, Online. Association for Computational Linguistics.

\bibitem[{Li et~al.(2024{\natexlab{a}})Li, Guo, Ma, He, Zhang, Rui, Ding, Li, Jian, Cheng, and Guo}]{li2024comparison}
JiaHang Li, ShuXia Guo, RuLin Ma, Jia He, XiangHui Zhang, DongSheng Rui, YuSong Ding, Yu~Li, LeYao Jian, Jing Cheng, and Heng Guo. 2024{\natexlab{a}}.
\newblock \href {https://doi.org/10.1186/s12874-024-01041-1} {Comparison of the effects of imputation methods for missing data in predictive modelling of cohort study datasets}.
\newblock \emph{BMC Medical Research Methodology}, 24(1):41.

\bibitem[{Li et~al.(2024{\natexlab{b}})Li, Zhou, Huang, Cheng, and Chen}]{li-etal-2024-eliciting}
Jiahuan Li, Hao Zhou, Shujian Huang, Shanbo Cheng, and Jiajun Chen. 2024{\natexlab{b}}.
\newblock \href {https://doi.org/10.1162/tacl_a_00655} {Eliciting the translation ability of large language models via multilingual finetuning with translation instructions}.
\newblock \emph{Transactions of the Association for Computational Linguistics}, 12:576--592.

\bibitem[{Lin et~al.(2019)Lin, Chen, Lee, Li, Zhang, Xia, Rijhwani, He, Zhang, Ma et~al.}]{lin2019choosing}
Yu-Hsiang Lin, Chian-Yu Chen, Jean Lee, Zirui Li, Yuyan Zhang, Mengzhou Xia, Shruti Rijhwani, Junxian He, Zhisong Zhang, Xuezhe Ma, et~al. 2019.
\newblock Choosing transfer languages for cross-lingual learning.
\newblock \emph{arXiv preprint arXiv:1905.12688}.

\bibitem[{Littell et~al.(2017)Littell, Mortensen, Lin, Kairis, Turner, and Levin}]{littell-etal-2017-uriel}
Patrick Littell, David~R. Mortensen, Ke~Lin, Katherine Kairis, Carlisle Turner, and Lori Levin. 2017.
\newblock \href {https://aclanthology.org/E17-2002} {{URIEL} and lang2vec: Representing languages as typological, geographical, and phylogenetic vectors}.
\newblock In \emph{Proceedings of the 15th Conference of the {E}uropean Chapter of the Association for Computational Linguistics: Volume 2, Short Papers}, pages 8--14, Valencia, Spain. Association for Computational Linguistics.

\bibitem[{Marsico et~al.(2018)Marsico, Flavier, Verkerk, and Moran}]{marsico-etal-2018-bdproto}
Egidio Marsico, Sebastien Flavier, Annemarie Verkerk, and Steven Moran. 2018.
\newblock \href {https://aclanthology.org/L18-1262} {{BDPROTO}: A database of phonological inventories from ancient and reconstructed languages}.
\newblock In \emph{Proceedings of the Eleventh International Conference on Language Resources and Evaluation ({LREC} 2018)}, Miyazaki, Japan. European Language Resources Association (ELRA).

\bibitem[{Mazumder et~al.(2010)Mazumder, Hastie, and Tibshirani}]{mazumder2010spectral}
Rahul Mazumder, Trevor Hastie, and Robert Tibshirani. 2010.
\newblock Spectral regularization algorithms for learning large incomplete matrices.
\newblock \emph{The Journal of Machine Learning Research}, 11:2287--2322.

\bibitem[{Michael et~al.(2015)Michael, Lev, Stark, Clem, and Chang}]{Michael_2013}
Michael, Lev, Tammy Stark, Emily Clem, and Will Chang. 2015.
\newblock \href {https://linguistics.berkeley.edu/saphon/en/cite.php} {South american phonological inventory database v2.1.0}.

\bibitem[{Mufwene(2013)}]{Mufwene_2013}
Salikoko~S. Mufwene. 2013.
\newblock \href {http://apics-online.info/contributions/58} {Atlas of pidgin and creole language structures online}.

\bibitem[{Nerbonne and Hinrichs(2006)}]{nerbonne-hinrichs-2006-linguistic}
John Nerbonne and Erhard Hinrichs. 2006.
\newblock \href {https://aclanthology.org/W06-1101} {Linguistic distances}.
\newblock In \emph{Proceedings of the Workshop on Linguistic Distances}, pages 1--6, Sydney, Australia. Association for Computational Linguistics.

\bibitem[{Ousidhoum et~al.(2024)Ousidhoum, Muhammad, Abdalla, Abdulmumin, Ahmad, Ahuja, Aji, Araujo, Ayele, Baswani, Beloucif, Biemann, Bourhim, Kock, Dekebo, Hourrane, Kanumolu, Madasu, Rutunda, Shrivastava, Solorio, Surange, Tilaye, Vishnubhotla, Winata, Yimam, and Mohammad}]{ousidhoum2024semrel2024collectionsemantictextual}
Nedjma Ousidhoum, Shamsuddeen~Hassan Muhammad, Mohamed Abdalla, Idris Abdulmumin, Ibrahim~Said Ahmad, Sanchit Ahuja, Alham~Fikri Aji, Vladimir Araujo, Abinew~Ali Ayele, Pavan Baswani, Meriem Beloucif, Chris Biemann, Sofia Bourhim, Christine~De Kock, Genet~Shanko Dekebo, Oumaima Hourrane, Gopichand Kanumolu, Lokesh Madasu, Samuel Rutunda, Manish Shrivastava, Thamar Solorio, Nirmal Surange, Hailegnaw~Getaneh Tilaye, Krishnapriya Vishnubhotla, Genta Winata, Seid~Muhie Yimam, and Saif~M. Mohammad. 2024.
\newblock \href {https://arxiv.org/abs/2402.08638} {Semrel2024: A collection of semantic textual relatedness datasets for 13 languages}.
\newblock \emph{Preprint}, arXiv:2402.08638.

\bibitem[{Papadimitriou and Jurafsky(2020)}]{papadimitriou-jurafsky-2020-learning}
Isabel Papadimitriou and Dan Jurafsky. 2020.
\newblock \href {https://doi.org/10.18653/v1/2020.emnlp-main.554} {{L}earning {M}usic {H}elps {Y}ou {R}ead: {U}sing transfer to study linguistic structure in language models}.
\newblock In \emph{Proceedings of the 2020 Conference on Empirical Methods in Natural Language Processing (EMNLP)}, pages 6829--6839, Online. Association for Computational Linguistics.

\bibitem[{Rey and Neuhäuser(2014)}]{Rey2014}
Denise Rey and Markus Neuhäuser. 2014.
\newblock \href {https://link.springer.com/referenceworkentry/10.1007/978-3-642-04898-2_616} {Wilcoxon-signed-rank test}.
\newblock In M.~M. Atkinson and W.~J. Piegorsch, editors, \emph{International Encyclopedia of Statistical Science}, pages 1658--1659. Springer.
\newblock First Online: 01 January 2014.

\bibitem[{Rubin(1987)}]{rubin1987multiple}
Donald~B. Rubin. 1987.
\newblock \href {https://doi.org/10.1002/9780470316696} {\emph{Multiple Imputation for Nonresponse in Surveys}}.
\newblock John Wiley \& Sons Inc., New York.

\bibitem[{Ruder et~al.(2021)Ruder, Constant, Botha, Siddhant, Firat, Fu, Liu, Hu, Garrette, Neubig, and Johnson}]{ruder-etal-2021-xtreme}
Sebastian Ruder, Noah Constant, Jan Botha, Aditya Siddhant, Orhan Firat, Jinlan Fu, Pengfei Liu, Junjie Hu, Dan Garrette, Graham Neubig, and Melvin Johnson. 2021.
\newblock \href {https://doi.org/10.18653/v1/2021.emnlp-main.802} {{XTREME}-{R}: Towards more challenging and nuanced multilingual evaluation}.
\newblock In \emph{Proceedings of the 2021 Conference on Empirical Methods in Natural Language Processing}, pages 10215--10245, Online and Punta Cana, Dominican Republic. Association for Computational Linguistics.

\bibitem[{Salati et~al.(2016)Salati, Falcoz, Decaluwe, Rocco, Raemdonck, Varela, Brunelli, and on~behalf of~the ESTS Database~Committee}]{Salati2016}
Michele Salati, Pierre-Emmanuel Falcoz, Herbert Decaluwe, Gaetano Rocco, Dirk~Van Raemdonck, Gonzalo Varela, Alessandro Brunelli, and on~behalf of~the ESTS Database~Committee. 2016.
\newblock \href {https://doi.org/10.1093/ejcts/ezv385} {The european thoracic data quality project: An aggregate data quality score to measure the quality of international multi-institutional databases}.
\newblock \emph{European Journal of Cardio-Thoracic Surgery}, 49(5):1470--1475.

\bibitem[{Samardzic et~al.(2024)Samardzic, Gutierrez, Bentz, Moran, and Pelloni}]{samardzic-etal-2024-measure}
Tanja Samardzic, Ximena Gutierrez, Christian Bentz, Steven Moran, and Olga Pelloni. 2024.
\newblock \href {https://doi.org/10.18653/v1/2024.findings-naacl.213} {A measure for transparent comparison of linguistic diversity in multilingual {NLP} data sets}.
\newblock In \emph{Findings of the Association for Computational Linguistics: NAACL 2024}, pages 3367--3382, Mexico City, Mexico. Association for Computational Linguistics.

\bibitem[{Skirg{\aa}rd et~al.(2023)Skirg{\aa}rd, Haynie, Blasi, Hammarstr{\"o}m, Collins, Latarche, Lesage, Weber, Witzlack-Makarevich, Passmore et~al.}]{skirgaard2023grambank}
Hedvig Skirg{\aa}rd, Hannah~J Haynie, Dami{\'a}n~E Blasi, Harald Hammarstr{\"o}m, Jeremy Collins, Jay~J Latarche, Jakob Lesage, Tobias Weber, Alena Witzlack-Makarevich, Sam Passmore, et~al. 2023.
\newblock Grambank reveals the importance of genealogical constraints on linguistic diversity and highlights the impact of language loss.
\newblock \emph{Science Advances}, 9(16):eadg6175.

\bibitem[{Srinivasan et~al.(2021)Srinivasan, Sitaram, Ganu, Dandapat, Bali, and Choudhury}]{srinivasan2021predicting}
Anirudh Srinivasan, Sunayana Sitaram, Tanuja Ganu, Sandipan Dandapat, Kalika Bali, and Monojit Choudhury. 2021.
\newblock Predicting the performance of multilingual nlp models.
\newblock \emph{arXiv preprint arXiv:2110.08875}.

\bibitem[{Toossi et~al.(2024)Toossi, Huai, Liu, Khiu, Do{\u{g}}ru{\"o}z, and Lee}]{toossi-etal-2024-reproducibility}
Hasti Toossi, Guo Huai, Jinyu Liu, Eric Khiu, A.~Seza Do{\u{g}}ru{\"o}z, and En-Shiun Lee. 2024.
\newblock \href {https://doi.org/10.18653/v1/2024.naacl-srw.25} {A reproducibility study on quantifying language similarity: The impact of missing values in the {URIEL} knowledge base}.
\newblock In \emph{Proceedings of the 2024 Conference of the North American Chapter of the Association for Computational Linguistics: Human Language Technologies (Volume 4: Student Research Workshop)}, pages 233--241, Mexico City, Mexico. Association for Computational Linguistics.

\bibitem[{{\"U}st{\"u}n et~al.(2020){\"U}st{\"u}n, Bisazza, Bouma, and van Noord}]{ustun-etal-2020-udapter}
Ahmet {\"U}st{\"u}n, Arianna Bisazza, Gosse Bouma, and Gertjan van Noord. 2020.
\newblock \href {https://doi.org/10.18653/v1/2020.emnlp-main.180} {{UD}apter: Language adaptation for truly {U}niversal {D}ependency parsing}.
\newblock In \emph{Proceedings of the 2020 Conference on Empirical Methods in Natural Language Processing (EMNLP)}, pages 2302--2315, Online. Association for Computational Linguistics.

\bibitem[{Xia et~al.(2020)Xia, Anastasopoulos, Xu, Yang, and Neubig}]{xia-etal-2020-predicting}
Mengzhou Xia, Antonios Anastasopoulos, Ruochen Xu, Yiming Yang, and Graham Neubig. 2020.
\newblock \href {https://doi.org/10.18653/v1/2020.acl-main.764} {Predicting performance for natural language processing tasks}.
\newblock In \emph{Proceedings of the 58th Annual Meeting of the Association for Computational Linguistics}, pages 8625--8646, Online. Association for Computational Linguistics.

\bibitem[{Zanon~Boito et~al.(2020)Zanon~Boito, Havard, Garnerin, Le~Ferrand, and Besacier}]{zanon-boito-etal-2020-mass}
Marcely Zanon~Boito, William Havard, Mahault Garnerin, {\'E}ric Le~Ferrand, and Laurent Besacier. 2020.
\newblock \href {https://aclanthology.org/2020.lrec-1.799} {{M}a{SS}: A large and clean multilingual corpus of sentence-aligned spoken utterances extracted from the {B}ible}.
\newblock In \emph{Proceedings of the Twelfth Language Resources and Evaluation Conference}, pages 6486--6493, Marseille, France. European Language Resources Association.

\bibitem[{Zhang and Toral(2019)}]{zhang-toral-2019-effect}
Mike Zhang and Antonio Toral. 2019.
\newblock \href {https://doi.org/10.18653/v1/W19-5208} {The effect of translationese in machine translation test sets}.
\newblock In \emph{Proceedings of the Fourth Conference on Machine Translation (Volume 1: Research Papers)}, pages 73--81, Florence, Italy. Association for Computational Linguistics.

\end{thebibliography}

\appendix

\section{Imputation Quality Test}

In this section, we describe more details about the experimental setup and results of the imputation quality test.

\subsection{Experimental Settings}
\label{sec:appendix-imputation}
Imputation on our data is always performed on aggregated data, either union or average. Before imputation, we allow users to fill in missing dialect data with parent language data, assuming typological similarity (e.g., filling American English with English data). This method is applied across all imputation quality tests and downstream tasks (see section \ref{sec:appendixE}).

We follow the imputation quality test from \citet{li2024comparison}, where $20\%$ of non-missing data is randomly removed, imputed, and the imputed values $y_{\mathsf{pred}}$ are compared to the original values $y_{\mathsf{true}}$ using a metric $\mathcal{L}$. Metrics $\mathcal{L}$ include accuracy, precision, recall, and F1 for union data (binary), and RMSE/MAE for average data (continuous). We optimize RMSE for average data to penalize outliers and F1 for union data due to class imbalance (more 0s). The method that optimizes $\mathcal{L}$ is selected as the best imputation approach.

We compare $k$-NN, MIDASpy, and SoftImpute against mean imputation as a baseline. Note that two additional steps are used in this methodology, depending on the imputation method. 

\paragraph{$k$-NN Imputation} We perform a hyperparameter search over $k = 3, 6, 9, 12, 15$ using 5-fold cross-validation to select $k$, optimizing $\mathcal{L}$, prior to comparing $k$-NN with other imputation methods.

\paragraph{MIDASpy Imputation} MIDASpy performs multiple imputations, pooling results based on Rubin’s rules \cite{rubin1987multiple}. We report the average $\mathcal{L}$ across 5 imputed datasets. The network is initialized with 3 layers and 256 units per layer, following \citet{lall2023efficient}.



\subsection{Detailed Results}
\label{sec:detailed-results}

A more granular analysis of imputation performance by feature type is shown in Tables \ref{tab:ftype-imputation-results-union} and \ref{tab:ftype-imputation-results-avg}. SoftImpute consistently performs best across all feature types. Notably, all methods perform far better on inventory and phonological data than on syntactic or phonological data. In particular, for SoftImpute, there is at least a $+7\%$ gap between 1) inventory and phonological data, and 2) syntactic and morphological data. This performance gap correlates with the proportion of missing values in each feature type. Specifically, we see that:
\begin{itemize}
    \item $85.73\%$ of morphological data is missing
    \item $91.25\%$ of syntactic data is missing 
    \item $77.06\%$ of inventory data is missing
    \item $81.88\%$ of phonological data is missing
\end{itemize}

Notably, syntactic and morphological data have the highest proportions of missing values. This suggests that the imputation algorithms perform better when when there are fewer missing values to impute.

\paragraph{Remark} The imputation quality $\gamma$ in Section \ref{sec:confidence score} can be the F1 score for union-aggregated data or $1 - \mathsf{RMSE}$ for average-aggregated data.

\begin{table}[!t]
    \centering
    \resizebox{\columnwidth}{!}{%
    \begin{tabular}{p{2cm}p{1.2cm}llll}
    \toprule
    \textbf{Imputation Method} & \textbf{Feature Type}& \textbf{Accuracy} & \textbf{Precision} & \textbf{Recall} & \textbf{F1} \\
    \midrule
    \multirow{4}{*}{Mean} & mor & 0.7428 & 0.6456 & 0.2914 & 0.4016 \\
    & syn & 0.7386 & 0.6491 & 0.5445 & 0.5922 \\
    & inv & 0.8975 & 0.8453 & 0.6970 & 0.7640 \\
    & pho & 0.8426 & 0.8174 & 0.8397 & 0.8284 \\
    \midrule
    \multirow{4}{*}{MIDASpy} & mor & 0.7962 & 0.6937 & 0.5582 & 0.6186 \\
    & syn & 0.8034 & 0.7580 & 0.6397 & 0.6939 \\
    & inv & 0.9101 & 0.8484 & 0.7577 & 0.8005 \\
    & pho & 0.8574 & 0.8492 & 0.8328 & 0.8409 \\
    \midrule
    \multirow{4}{*}{$k$(9)-NN} & mor & 0.8388 & 0.7611 & \textbf{0.6644} & \textbf{0.7095} \\
    & syn & 0.8263 & 0.7830 & \textbf{0.6942} & 0.7359 \\
    & inv & 0.9301 & 0.8806 & 0.8173 & 0.8477 \\
    & pho & 0.8566 & 0.8547 & 0.8231 & 0.8386 \\
    \midrule
    \multirow{4}{*}{SoftImpute} & mor & \textbf{0.8485} & \textbf{0.8259} & 0.6188 & 0.7075 \\
    & syn & \textbf{0.8475} & \textbf{0.8518} & 0.6808 & \textbf{0.7568} \\
    & inv & \textbf{0.9444} & \textbf{0.9249} & \textbf{0.8342} & \textbf{0.8772} \\
    & pho & \textbf{0.9376} & \textbf{0.9657} & \textbf{0.8938} & \textbf{0.9284} \\
    \bottomrule
    \end{tabular}
    }
    \caption{Summary of feature type (syntactic (syn), phonological (pho), inventory (inv), and morphological (mor)) analysis of imputation quality test over union-aggregated data. \textbf{Bolded} entries indicate the best results in each category.}
    \label{tab:ftype-imputation-results-union}
\end{table}

\begin{table}[htp]
    \centering
    \resizebox{\columnwidth}{!}{%
    \begin{tabular}{llll}
    \toprule
    \textbf{Imputation Method} & \textbf{Feature Type} & \textbf{RMSE} & \textbf{MAE} \\
    \midrule
    \multirow{4}{*}{Mean} & mor & 0.4103 & 0.3370 \\
    & syn & 0.4089 & 0.3352 \\
    & inv & 0.2636 & 0.1428 \\
    & pho & 0.3464 & 0.2442 \\
    \midrule
    \multirow{4}{*}{MIDASpy} & mor & 0.3771 & 0.2814 \\
    & syn & 0.3688 & 0.2695 \\
    & inv & 0.2516 & 0.1278 \\
    & pho & 0.3227 & 0.2120 \\
    \midrule
    \multirow{4}{*}{$k$(15)-NN} & mor & 0.3471 & \textbf{0.2315} \\
    & syn & 0.3482 & \textbf{0.2339} \\
    & inv & 0.2234 & \textbf{0.0965} \\
    & pho & 0.3219 & 0.1857 \\
    \midrule
    \multirow{4}{*}{SoftImpute} & mor & \textbf{0.3398} & 0.2485 \\
    & syn &\textbf{0.3344} & 0.2451 \\
    & inv & \textbf{0.1987} & 0.1024 \\
    & pho & \textbf{0.2296} & \textbf{0.1447} \\
    \bottomrule
    \end{tabular}
    }
    \caption{Summary of feature type (syntactic (syn), phonological (pho), inventory (inv), and morphological (mor)) analysis of imputation quality test over average-aggregated data. \textbf{Bolded} entries indicate the best results in each category.}
    \label{tab:ftype-imputation-results-avg}
\end{table}


\section{Downstream Tasks}

In this section, we describe more details about the experimental setup and results for downstream tasks.

\subsection{Experimental Settings}
\label{sec:appendixE}

\begin{table}[!ht]
    \centering
    \begin{tabular}{lll}
    \toprule
    \textbf{Language} & \textbf{\uriel{} code} & \textbf{\uriel{+} code} \\
    \midrule
    Albanian & alb & alba1267 \\
    Arabic & ara & stan1318 \\
    Azerbaijani & aze & nort2697 \\
    Chinese & zho & mand1415 \\
    Estonian & ekk & esto1258 \\
    Malay & msa & stan1306 \\
    Oromo & orm & east2652 \\
    Persian & fas & west2369 \\
    Swahili & swa & swah1253 \\
    \bottomrule
    \end{tabular}%
    \caption{Languages used in downstream task experiments with ISO 639-3 codes in \uriel{} but without equivalent glottocodes in \uriel{+}, and the glottocodes that were used as replacements.}
    \label{tab:langs-downstream}
\end{table}

To run \langrank{} and \proxylm{} experiments with \uriel{+} distances, we simply replaced the \uriel{} distances with the new \uriel{+} distances.

While most languages used for downstream tasks had ISO 639-3 codes with corresponding glottocodes, some did not, often because the ISO 639-3 codes were outdated. In these cases, we assigned the glottocode of the most appropriate language from \uriel{+}. The languages without glottocodes and their replacements are listed in Table \ref{tab:langs-downstream}. 

No GPU was required for \langrank{} and \proxylm{} experiments. All experiments for \lingualchemy{} were run on two Tesla V100 32GB GPUs. 

\paragraph{Experimental Settings for \langrank{}}
We could not completely replicate the original \langrank{} baselines \cite{lin2019choosing} due to unclear parameter specifications. We attempted to keep most of the experimental setup the same as the original paper, except for the following:
\begin{itemize}
    \item When assigning relevance to languages that have the same BLEU or accuracy score, the lowest rank in the group is used, as it made the most sense and produced good results.
    \item For calculation of Normalized Discounted Cumulative Gain (NDCG), we attempted to use the original paper's formula where the Discounted Cumulative Gain is defined as
    \[\text{DCG}@p = \sum_{i=1}^p \frac{2^\gamma_i - 1}{\log_2(i+1)}.\]
    However, an alternative formulation of DCG, defined as 
    \[\text{DCG}@p = \sum_{i=1}^p \frac{\gamma_i}{\log_2(i+1)},\]
    was chosen instead, as it produced baseline results much closer to the original paper.
    
\end{itemize}

\paragraph{Experimental Settings for \lingualchemy{}}
We could not find the source code for the vectors for the MasakhaNews \cite{adelani-etal-2023-masakhanews} news classification dataset and the SemRel2024 \cite{ousidhoum2024semrel2024collectionsemantictextual} semantic relatedness dataset, as well as a pipeline for SemRel2024. To address this, we created vectors with both \uriel{} and \uriel{+} feature data for these datasets and constructed a pipeline for SemRel2024, using the approach applied to the MASSIVE \cite{fitzgerald-etal-2023-massive} intent classification dataset and MasakhaNews.

Only new syntactic vectors needed to be created, as the geography vectors for all languages in the \lingualchemy{} datasets remained unchanged. Despite their names, the syntax\_knn and syntax\_average vectors used SoftImpute with union and average aggregation, respectively, rather than $k$-NN, since SoftImpute was employed for imputation in all other downstream task experiments.

A 10$\times$ \uriel{} loss scaling factor was used as it provided the best results in \citealp{adilazuarda2024lingualchemy}. 

\paragraph{Experimental Settings for \proxylm{}}
The proxy regressor in \proxylm{} had two different datasets: an English-centric dataset borrowed from the MT560 dataset \cite{gowda-etal-2021-many} and a Many-to-many languages dataset borrowed from the NUSA dataset \cite{cahyawijaya-etal-2023-nusawrites}. The Many-to-many languages dataset contained the language Batak which does not have an ISO 639-3 code. \citealp{anugraha2024proxylm} used the code “bhp” for Bima, a Batak family member, which has a glottocode equivalent: “bima1247”.
We used the Ensemble model, which performed best for \proxylm{}. The hyperparameters of the proxy regressor were the same as in \citealp{anugraha2024proxylm}. 

\subsection{Detailed Results}\label{sec:appendix-performance}

Tables~\ref{tab:results-langrank1}, ~\ref{tab:results-lingualchemy}, and ~\ref{tab:results-proxylm}  show the results for using \uriel{+} on \langrank{}, \lingualchemy{}, and \proxylm{} respectively.

\begin{table}[!t]
    \centering
    \begin{tabular}{lll}
    \toprule
    \textbf{Language} & \textbf{\uriel{} code} & \textbf{\uriel{+} code} \\ 
    \midrule
    Sambu & emp & nort2972 \\
    Cayapa & cbi & chac1249 \\
    Paya & pay & pech1241 \\
    Bintucua & arh & arhu1242 \\
    Cágaba & kog & cogu1240 \\
    Ulua & sum & sumu1234 \\
    Paez & pbb & paez1247 \\
    Sumo & sum & sumu1234 \\
    Cuna & cuk & sanb1242 \\
    Boruca & brn & boru1252 \\
    Muisca & chb & chib1270 \\
    Huaunana & noa & woun1238 \\
    Misquito & miq & misk1235 \\
    Quiche & quc & kich1262 \\
    Lenca & len & lenc1239 \\
    Xinca & xin & xinc1237 \\
    Camsa & kbh & cams1241 \\
    Cofan & con & cofa1242 \\
    Colorado & cof & colo1256 \\
    \bottomrule
    \end{tabular}%
    \caption{Languages used in the distance accuracy case study with corresponding identifiers. The table includes \textbf{\uriel{} code} using \textbf{ISO 639-3} identifiers and \textbf{\uriel{+} code} using \textbf{glottocode} identifiers.}
    \label{tab:langs-casestudy}
\end{table}

\begin{table*}[!th]
\centering
\resizebox{0.99\textwidth}{!}{
    \begin{tabular}{rrrrrrrrr}
    \toprule   
    \multicolumn{1}{c}{\multirow{2}{*}{\textbf{Method}}} & \multicolumn{2}{c}{\textbf{Machine Translation}} & \multicolumn{2}{c}{\textbf{Entity Linking}} & \multicolumn{2}{c}{\textbf{POS Tagging}} & \multicolumn{2}{c}{\textbf{Dependency Parsing}}\\
    &
    \multicolumn{1}{c}{\textbf{\uriel{}}} & \multicolumn{1}{c}{\textbf{\methodname{}}} & 
    \multicolumn{1}{c}{\textbf{\uriel{}}} & \multicolumn{1}{c}{\textbf{\methodname{}}} & 
    \multicolumn{1}{c}{\textbf{\uriel{}}} & \multicolumn{1}{c}{\textbf{\methodname{}}} & 
    \multicolumn{1}{c}{\textbf{\uriel{}}} & \multicolumn{1}{c}{\textbf{\methodname{}}} \\
    \midrule
    \langrank{} (lang feats) & 35.2 & 35.4 (\textcolor{darkblue}{$\uparrow 0.20$}) & 64.7 & 66.4 (\textcolor{darkblue}{$\uparrow 1.70$}) & 18.3 & 20.4  (\textcolor{darkblue}{$\uparrow 2.10$}) & 73.2 & 70.4 (\textcolor{darkred}{$\downarrow 2.8$})\\
    \langrank{} (all) & 49.9 & 55.5 (\textcolor{darkblue}{$\uparrow 5.60$}) & 68.3 & 67.2 (\textcolor{darkred}{$\downarrow 1.10$}) & 27.5 & 28.8  (\textcolor{darkblue}{$\uparrow 1.30$}) & 70.2 & 75.3 (\textcolor{darkblue}{$\uparrow 5.10$})\\
    \midrule
    \multicolumn{1}{c}{\textbf{Avg. \langrank{}}} & 42.55 & 45.45 (\textcolor{darkblue}{$\uparrow 2.90$}) &66.50 & 66.80 (\textcolor{darkblue}{$\uparrow 0.30$}) & 22.9 & 24.6 (\textcolor{darkblue}{$\uparrow 1.70$}) &71.7 & 72.85 (\textcolor{darkblue}{$\uparrow 1.15$})\\
    \bottomrule
    \end{tabular}
}
\caption{The results for two \langrank{} model~\cite{lin2019choosing} on predicting cross-lingual transfer measured using average NDCG@3 multiplied by $100$ (\textbf{higher is better}). \langrank{} (lang feats) considers only language vectors, while \langrank{} (all) denotes \langrank{} with language vectors and additional dataset-dependent features such as size and type-token ratio.}
\label{tab:results-langrank1}
\end{table*}

\begin{table*}[!th]
\centering
\resizebox{0.99\textwidth}{!}{
    \begin{tabular}{lrrrrrrr}
    \toprule   
    \multirow{2}{*}{\textbf{Model}} & \multicolumn{1}{c}{\multirow{2}{*}{\textbf{Feature Type}}} & \multicolumn{2}{c}{\textbf{MASSIVE}} & \multicolumn{2}{c}{\textbf{MasakhaNews}} & \multicolumn{2}{c}{\textbf{SemRel2024}}\\
    & & \multicolumn{1}{c}{\textbf{\uriel{}}} & \multicolumn{1}{c}{\textbf{\methodname{}}} & \multicolumn{1}{c}{\textbf{\uriel{}}} & \multicolumn{1}{c}{\textbf{\methodname{}}} & \multicolumn{1}{c}{\textbf{\uriel{}}} & 
    \multicolumn{1}{c}{\textbf{\methodname{}}} \\
    \midrule
    \multirow{6}{*}{mBERT} & syntax\_avg & 66.01 & 65.71 (\textcolor{darkred}{$\downarrow 0.30$}) & 70.96 & 71.52 (\textcolor{darkblue}{$\uparrow 0.56$}) & 0.005 & 0.009 (\textcolor{darkblue}{$\uparrow 0.004$}) \\
    & syntax\_avg+geo & 65.59 & 65.74 (\textcolor{darkblue}{$\uparrow 0.15$}) & 70.77 & 70.75 (\textcolor{darkred}{$\downarrow 0.02$}) & 0.01 & 0.008 (\textcolor{darkred}{$\downarrow 0.002$}) \\
    & syntax\_knn & 65.83 & 65.85 (\textcolor{darkblue}{$\uparrow 0.02$}) & 71.08 & 72.07 (\textcolor{darkblue}{$\uparrow 0.99$}) & 0.014 & 0.012 (\textcolor{darkred}{$\downarrow 0.002$}) \\
    & syntax\_knn+geo & 65.45 & 66.00 (\textcolor{darkblue}{$\uparrow 0.55$}) & 70.91 & 71.36 (\textcolor{darkblue}{$\uparrow 0.45$}) & 0.008 & 0.013 (\textcolor{darkblue}{$\uparrow 0.005$}) \\
    & syntax\_knn+syntax\_avg & 65.45 & 66.25 (\textcolor{darkblue}{$\uparrow 0.80$}) & 71.22 & 70.52 (\textcolor{darkred}{$\downarrow 0.70$}) & 0.013 & 0.015 (\textcolor{darkblue}{$\uparrow 0.002$}) \\
    & syntax\_knn+syntax\_avg+geo & 65.59 & 65.82 (\textcolor{darkblue}{$\uparrow 0.23$}) & 70.99 & 71.13 (\textcolor{darkblue}{$\uparrow 0.14$}) & 0.01 & 0.012 (\textcolor{darkblue}{$\uparrow 0.002$}) \\
    \midrule
    \multirow{6}{*}{XLM-R} & syntax\_avg & 80.51 & 80.47 (\textcolor{darkred}{$\downarrow 0.04$}) & 81.31 & 80.91 (\textcolor{darkred}{$\downarrow 0.40$}) & 0.007 & 0.012 (\textcolor{darkblue}{$\uparrow 0.005$}) \\
    & syntax\_avg+geo & 80.20 & 80.35 (\textcolor{darkblue}{$\uparrow 0.15$}) & 81.58 & 81.71 (\textcolor{darkblue}{$\uparrow 0.13$}) & 0.015 & 0.013 (\textcolor{darkred}{$\downarrow 0.002$}) \\
    & syntax\_knn & 80.11 & 80.53 (\textcolor{darkblue}{$\uparrow 0.42$}) & 81.18 & 81.40 (\textcolor{darkblue}{$\uparrow 0.22$}) & 0.017 & 0.013 (\textcolor{darkred}{$\downarrow 0.004$}) \\
    & syntax\_knn+geo & 80.00 & 80.33 (\textcolor{darkblue}{$\uparrow 0.33$}) & 81.32 & 81.90 (\textcolor{darkblue}{$\uparrow 0.58$}) & 0.005 & 0.007 (\textcolor{darkblue}{$\uparrow 0.002$}) \\
    & syntax\_knn+syntax\_avg & 80.47 & 80.19 (\textcolor{darkred}{$\downarrow 0.28$}) & 80.63 & 82.00 (\textcolor{darkblue}{$\uparrow 1.37$}) & 0.012 & 0.014 (\textcolor{darkblue}{$\uparrow 0.002$}) \\
    & syntax\_knn+syntax\_avg+geo & 80.17 & 80.39 (\textcolor{darkblue}{$\uparrow 0.22$}) & 81.58 & 81.89 (\textcolor{darkblue}{$\uparrow 0.31$}) & 0.008 & 0.010 (\textcolor{darkblue}{$\uparrow 0.002$}) \\
    \midrule
    \multicolumn{2}{c}{\textbf{Avg.}} & 72.95 & 73.16 (\textcolor{darkblue}{$\uparrow 0.21$}) & 76.13 & 76.43 (\textcolor{darkblue}{$\uparrow 0.30$}) & 0.008 & 0.012 (\textcolor{darkblue}{$\uparrow 0.004$}) \\
    \bottomrule
    \end{tabular}
}
\caption{The results for \lingualchemy{}~\cite{adilazuarda2024lingualchemy} using \uriel{} loss scale of 10 obtained by averaging the accuracy for MASSIVE and MasakhaNews datasets and the Pearson correlation for SemRel2024 dataset across all languages under different benchmarks (\textbf{higher is better}). \textbf{\uriel{}} denotes \lingualchemy{} with \uriel{} vectors, while \textbf{\methodname{}} denotes \lingualchemy{} with \methodname{} vectors.}
\label{tab:results-lingualchemy}
\end{table*}

\begin{table*}[!th]
\centering
\resizebox{0.99\textwidth}{!}{
    \begin{tabular}{lrrrrrrr}
    \toprule   
    \multirow{2}{*}{\textbf{Dataset}} & \multicolumn{1}{c}{\multirow{2}{*}{\textbf{Experimental Setting}}} & \multicolumn{2}{c}{\textbf{M2M100}} & \multicolumn{2}{c}{\textbf{NLLB}} \\
    & & \multicolumn{1}{c}{\textbf{URIEL}} & \multicolumn{1}{c}{\textbf{\methodname{}}} & \multicolumn{1}{c}{\textbf{URIEL}} & \multicolumn{1}{c}{\textbf{\methodname{}}} & \\
    \midrule
    \multirow{3}{*}{\textbf{English Centric}} & Random & 3.64 \scriptsize $\pm$ 0.19 & 3.62 \scriptsize $\pm$ 0.18 \normalsize (\textcolor{darkblue}{$\downarrow 0.02$}) & 3.80 \scriptsize $\pm$ 0.37 & 3.79 \scriptsize $\pm$ 0.39 \normalsize (\textcolor{darkblue}{$\downarrow 0.01$}) \\
    & LOLO & 3.90 \scriptsize $\pm$ 0.22 & 3.84 \scriptsize $\pm$ 0.22 \normalsize (\textcolor{darkblue}{$\downarrow 0.06$}) & 4.14 \scriptsize $\pm$ 0.23 & 4.10 \scriptsize $\pm$ 0.23 \normalsize (\textcolor{darkblue}{$\downarrow 0.04$}) \\
    & Unseen & 4.35 \scriptsize $\pm$ 0.25 & 4.08 \scriptsize $\pm$ 0.25 \normalsize (\textcolor{darkblue}{$\downarrow 0.27$}) & \multicolumn{1}{c}{NA} & \multicolumn{1}{c}{NA} \\
    \midrule
    \multirow{2}{*}{\textbf{Many-to-Many}} & Random & 2.47 \scriptsize $\pm$ 0.35 & 2.36 \scriptsize $\pm$ 0.29 \normalsize (\textcolor{darkblue}{$\downarrow 0.11$}) & 1.76 \scriptsize $\pm$ 0.42 & 1.49 \scriptsize $\pm$ 0.32 \normalsize (\textcolor{darkblue}{$\downarrow 0.27$}) \\
    & LOLO & 3.64 \scriptsize $\pm$ 0.24 & 3.67 \scriptsize $\pm$ 0.24 \normalsize (\textcolor{darkred}{$\uparrow 0.03$}) & 3.67 \scriptsize $\pm$ 0.18 & 3.79 \scriptsize $\pm$ 0.26 \normalsize (\textcolor{darkred}{$\uparrow 0.12$}) \\
    \midrule
    \multicolumn{2}{c}{\textbf{Avg.}} & 3.60 & 3.51 (\textcolor{darkblue}{$\downarrow 0.09$}) & 3.34 & 3.29 (\textcolor{darkblue}{$\downarrow 0.05$}) \\
    \bottomrule
    \end{tabular}
}
\caption{The results for \proxylm{}~\cite{anugraha2024proxylm} using XGBoost Ensemble in average RMSE $\pm$ standard deviation under different datasets and settings (\textbf{lower is better}). \textbf{URIEL} denotes \proxylm{} with URIEL as its language features, while \textbf{\methodname{}} denotes \proxylm{} with \methodname{} as its language features.}
\label{tab:results-proxylm}
\end{table*}

\end{document}